\documentclass{article} % For LaTeX2e
\usepackage{iclr2018_conference,times}
\usepackage{hyperref}
\usepackage{url}
\usepackage{algorithm}
\usepackage{algorithmic}
\usepackage{bm}
\usepackage{amsfonts}
\usepackage{nicefrac}
\usepackage{color}  
\usepackage{graphics}
\usepackage{mathtools}
\usepackage{subfigure}
\usepackage{wrapfig}
\usepackage{booktabs}
\usepackage{tikz}
\usepackage{authblk}
\newcommand*{\affaddr}[1]{#1} % No op here. Customize it for different styles.
\newcommand*{\affmark}[1][*]{\textsuperscript{#1}}
\newcommand*{\email}[1]{\texttt{#1}}
\title{Adversarial Dropout Regularization}
\author{%
  Kuniaki Saito\affmark[1], Yoshitaka Ushiku\affmark[1], Tatsuya Harada\affmark[1,2], and Kate Saenko\affmark[3]\\
  \affaddr{\affmark[1]The University of Tokyo}, \affaddr{\affmark[2]RIKEN}, \affaddr{\affmark[3]Boston University}\\
  \email{\{k-saito,ushiku,harada\}@mi.t.u-tokyo.ac.jp},   \email{saenko@bu.edu}\\
  }
\iclrfinalcopy % Uncomment for camera-ready version, but NOT for submission.

\begin{document}

\maketitle

\begin{abstract}
%Changed first sentence to summarize contribution right away
We present a method for transferring neural representations from label-rich source domains to unlabeled target domains. Recent adversarial methods proposed for this task learn to align features across domains by fooling a special domain critic network. However, a drawback of this approach is that the critic simply labels the generated features as in-domain or not, without considering the boundaries between classes. This can lead to ambiguous features being generated near class boundaries, reducing target classification accuracy. We propose a novel approach, Adversarial Dropout Regularization (ADR), to encourage the generator to output more discriminative features for the target domain. Our key idea is to replace the critic with one that detects non-discriminative features, using dropout on the classifier network. The generator then learns to avoid these areas of the feature space and thus creates better features. We apply our ADR approach to the problem of unsupervised domain adaptation for image classification and semantic segmentation tasks, and demonstrate significant improvement over the state of the art. We also show that our approach can be used to train Generative Adversarial Networks for semi-supervised learning. 
%  Adversarial learning methods have been used for aligning distributions between different domains to transfer the knowledge. To the best of our knowledge, such methods try to minimize divergence between features without considering the relationship with decision boundary. In order to extract discriminative features for target domain, we propose a new adversarial learning method which accounts for the information of decision boundary. We present a novel method utilizing dropout. Our method regularize the feature extraction networks to learn discriminative information. We train classifier networks to be sensitive to the noise caused by dropout. Then, the feature extraction networks are trained to be less sensitive to the noise, thus achieve regularization of the feature extraction networks. Our proposed method is directly applicable to domain adaptation in semantic segmentation, and outperformed other methods in many settings. Further, our method is show to be effective in training Generative Adversarial Networks.
\end{abstract}

\section{Introduction}
\vspace{-3mm}        
%1, Transferring the knowledge from source to target domain is an important problem.
%
%2, The core challenges are that source and target samples have different characteristics, and how to transfer the knowledge between different domains.

\begin{figure}[t]
  \begin{center}
  \includegraphics[width=\hsize]{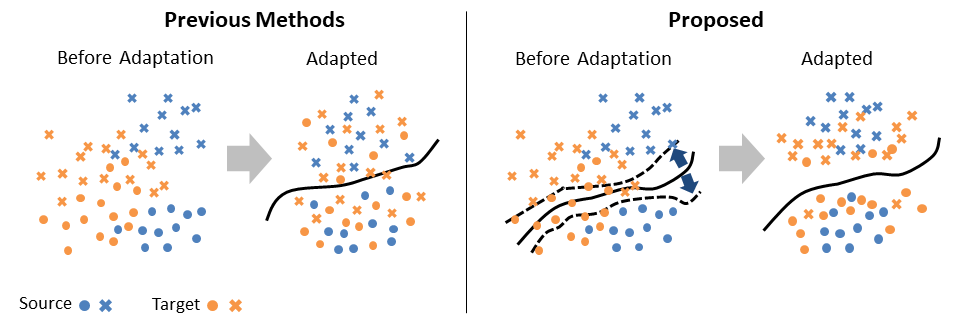}
  \end{center}
 \vspace{-0.1in}
  \caption{{\small (a-d) An illustration of a deep model trained on simulated source training data failing to segment a real target domain image: (a) shows the target image, (b) is the ground truth segmentation into semantic categories (car, road, etc), (d) is the output of the unadapted source models, (e) improved segmentation obtained by our proposed ADR method. (e) Previous distribution matching methods do not consider the source decision boundary when aligning source and target feature points. (f) We propose to use the boundary information to achieve low-density separation of aligned points.}}
  \label{fig:intro}
  \end{figure}

Transferring knowledge learned by deep neural networks on label-rich domains to new target domains is a challenging problem, especially when the source and target input distributions have different characteristics. For example, while simulated driving images rendered by games provide a rich source of labeled data for semantic segmentation, deep models trained on such source data do not transfer well to real target domains (Fig.~\ref{fig:intro} (a-d)). When target-domain labels are unavailable for fine-tuning, unsupervised domain adaptation must be applied to improve the source model. 

Recent methods for unsupervised domain adaptation attempt to reduce the discrepancy between the source and target features via adversarial learning~(\cite{tzeng2014deep,ganin2014unsupervised}). They divide the base network into a feature encoder $G$ and classifier $C$, and add a separate domain classifier (critic) network $D$. The critic takes the features generated by $G$ and labels them as either source- or target-domain. The encoder $E$ is then trained with an additional adversarial loss that maximizes $D$'s mistakes and thus aligns features across domains.

However, a major drawback of this approach is that the critic simply predicts the domain label of the generated point and does not consider category information. Thus the generator may create features that look like they came from the right domain, but are not discriminative. In particular, it can generate points close to class boundaries, as shown in Fig.~\ref{fig:intro}(e), which are likely to be mis-classified by the source model. We argue that to achieve good performance on the target data, the adaptation model must take the decision boundaries between classes into account while aligning features across domains (Fig.~\ref{fig:intro}(f)). The problem is, it is not clear how this can be accomplished without labels on target data.

In this paper, we propose a novel adversarial alignment technique that overcomes the above limitation and preserves class boundaries.
We make the following observation: if the critic could detect points near the decision boundary, then the generator would have to avoid these areas of the feature space in order to fool the critic. Thus the critic would force the generator to create more discriminative features. How can we obtain such a critic? If we alter the boundary of the classifier $C$ slightly and measure the change in the posterior class probability $p(y|x)$, where $y$ and $x$ denote class and input respectively, then samples near the decision boundary are likely to have the largest change. In fact, this posterior discrepancy is inversely proportional to the distance from the class boundary. We thus propose to maximize this posterior discrepancy to turn $C$ into a critic sensitive to non-discriminative points. We call this technique \textit{Adversarial Dropout Regularization}. Here, dropout is not used in the standard way, which is to regularize the main classifier and make it insensitive to noise. Instead, we use dropout in an adversarial way, to transform the classifier into a critic sensitive to noise. Compared to previous adversarial feature alignment methods, where the distributions $p(x)$ are aligned globally, our method aligns target features away from decision boundaries, as illustrated in Fig.\ref{fig:intro}(f). 
%change figure

Our ADR approach has several benefits.
First, we train the generator $G$ with feedback from the classifier $C$, in contrast to existing methods, which use an unrelated critic $D$.
Second, our method is general and straightforward to apply to a variety of domain adaptation problems, such as classification and semantic segmentation.
Finally, since ADR is trained to align distributions, it is also applicable to semi-supervised learning and training of generative models, such as Generative Adversarial Networks (GANs)~(\cite{goodfellow2014generative}). Through extensive experiments, we demonstrate the benefit of ADR over existing domain adaptation approaches, achieving state-of-the-art results in difficult domain shifts. We also show an application to semi-supervised learning using GANs.

\section{Related Work}\label{sec:related}
\vspace{-3mm}        

\noindent {\bf Domain Adaptation.}
%In domain adaptation (DA), we aim to transfer the knowledge learned on a labeled source domain to a new target domain which lacks labeled samples~\citep{saenko_eccv10}. Unsupervised domain adaptation (UDA), where we do not have any labeled samples for target domain, is an especially challenging and important problem. 
Recent unsupervised domain adaptation (UDA) methods for visual data aim to align the feature distributions of the source and target domains~(\cite{sun2016return,sun2016deep,tzeng2014deep,ganin2016domain,long2015learning,yan2017mind,long2016deep}). Such methods are motivated by theoretical results stating that minimizing the divergence between domains will lower the upper bound of the error on target domain~(\cite{ben2010theory}). 
Many works in deep learning utilize the technique of distribution matching in hidden layers of a network such as a CNN~(\cite{tzeng2014deep,ganin2016domain,long2015learning}). However, they measure the domain divergence based on the hidden features of the network without considering the relationship between its decision boundary and the target features, as we do.
%already argued this above
%However, this relationship should be utilized for successful %adaptation and generalization. 

\noindent {\bf Low-density Separation.} Many semi-supervised learning (SSL) methods utilize the relationship between the decision boundary and unlabeled samples, a technique called low-density separation~(\cite{chapelle2005semi,joachims1999transductive}). By placing the boundary in the area where the unlabeled samples are sparse, these models aim to obtain discriminative representations.
Our method aims to achieve low-density separation for deep domain adaptation and is related to entropy minimization for semi-supervised learning~(\cite{grandvalet2005semi}). (\cite{long2016unsupervised}) used entropy minimization in their approach to directly measure how far samples are from a decision boundary by calculating entropy of the classifier's output. On the other hand, our method tries to achieve low-density separation by slightly moving the boundary and detecting target samples sensitive to the movement. As long as target samples features are robust to the movement, they will be allowed to exist relatively nearby the boundary compared to source samples, as Fig.~\ref{fig:intro} shows. 
%\textcolor{red}{KS:I did not really understand this point}
% \textcolor{blue}{Kuniaki: I think this part is not necessary too. I just wanted to say the same thing explained in the sentence before this line.}
% This characteristic is advantageous in DA as it is often difficult to separate target samples into each class correctly as source samples are separated.

%\textcolor{blue}{add sentences on the new baseline method}
%\textcolor{red}{say whether we compare to Long2016?}
%In the experimental section, we compare our method with previous distribution matching based methods, as well as
In (\cite{long2016unsupervised}) entropy minimization is just a part of the overall approach. To compare our ADR approach to entropy minimization directly, we use a new baseline method. To our knowledge, though this method has not been proposed by any previous works, it is easily achieved by modifying a method proposed by~(\cite{springenberg2015unsupervised}). The generator tries to minimize the entropy of the target samples whereas the critic tries to maximize it. The entropy is directly measured by the output of the classifier. This baseline is similar to our approach in that the goal of the method is to achieve low-density separation. 
%We describe this baseline in the method section.
%\textcolor{red}{I would just point to method here, then point to appendix from there} space limitations, we provide a detailed explanation in the appendix. The explanation will be easier to understand after reading our method section.

\noindent {\bf Dropout.} Dropout is a method that prevents the networks from overfitting~(\cite{srivastava2014dropout}) by randomly dropping units from the neural network during training. Effectively, dropout samples from an exponential number of different thinned networks at training time, which prevents units from co-adapting too much. At test time, predictions are obtained by using the outputs of all neurons. If the thinned networks are able to classify the samples accurately, the full network will as well. In other words, dropout encourages the network to be robust to noise. In our work, we use dropout to regularize the feature generation network $G$, but in an adversarial way. We train the critic $C$ to be sensitive to the noise caused by dropout and use $C$ to regularize $G$ so that it generates noise-robust features. To our knowledge, this use of dropout is completely different from existing methods.

\section{Method}\label{sec:method}
\vspace{-3mm}        
% \textcolor{red}{say what the input to the algorithm is -- what data? talk about pre-training C on data? and G?}
% \textcolor{blue}{add explanation here. I add explanations on notation and settings here.}
We assume that we have access to a labeled source image $\mathbf{x_{s}}$ and a corresponding label $y_{s}$ drawn from a set of labeled source images \{$X_{s},Y_{s}$\},
as well as an unlabeled target image $\mathbf{x_{t}}$ drawn from unlabeled target images $X_{t}$. 
We train a feature generation network $G$, which takes inputs $\mathbf{x_{s}}$ or $\mathbf{x_{t}}$, and a network $C$ that acts as both the main classifier and the critic. $C$ takes features from $G$ and classifies them into $K$ classes, predicting a $K$-dimensional vector of logits \{$l_{1},l_{2},l_{3}...l_{K}$\}. The logits are then converted to class probabilities by applying the softmax function. Namely, the probability that $\mathbf{x}$ is classified into class $j$ is denoted by $p(y=j|{\mathbf{x}})=\frac{exp(l_{j})}{\sum_{k=1}^{K}exp(l_{k})}$. We use the notation $p(\mathbf{y}|\mathbf{x})$ to denote the $K$-dimensional probabilistic output for input $\mathbf{x}$.

The weights of $G$ can be initialized either by pre-training on some auxiliary dataset (e.g., Imagenet), or with random weights. $C$ uses random initialization. 
%The following explanation is common both for when using pretrained model or training a model from scratch.
When acting as critic, $C$ must detect the feature encodings of target samples near the decision boundary, while $G$ must avoid generating features near that area to fool the critic. To detect the samples near the decision boundary, we propose to slightly perturb the boundary and measure the change in the posterior class probability $p(\mathbf{y}|\mathbf{x})$. The critic network $C$ is then trained to increase the change while the feature generation network $G$ is trained to decrease it. Through this adversarial training, $G$ learns to generate target features far away from the decision boundary to decrease the change in posterior, as samples near the decision boundary are likely to have the largest change. 
In this section, we show how we utilize dropout to perturb the boundary in the critic and measure sensitivity. We then show how to construct the training procedure to ensure discriminativeness for source samples, without which the decision boundary would be meaningless. Finally, we give some intuition for our method and improve it based on this insight.
As the training procedure used for GAN is slightly different from the one used in domain adaptation experiments, we describe it in the corresponding experimental section. 

\subsection{Classifier Selection via Dropout}
\vspace{-3mm}        
%Consider a standard classifier $C$ for classifying sample $x$ into $K$ class. $C$ outputs a $K$-dimensional vector  per one sample. Then, we can obtain class probabilities by applying softmax function. Without any notation, we consider that $C$ outputs $K$-dimensional vector indicating class probabilities. \textcolor{red}{but we use notation later, no? may as well define it here}

Consider the standard training of a neural network using dropout. For every sample within a mini-batch, each node of the network is removed with some probability, effectively selecting a different classifier for every sample during training. We harness this idea in a very simple way.

\begin{figure}[t]
  \begin{center}
  \includegraphics[width=\hsize]{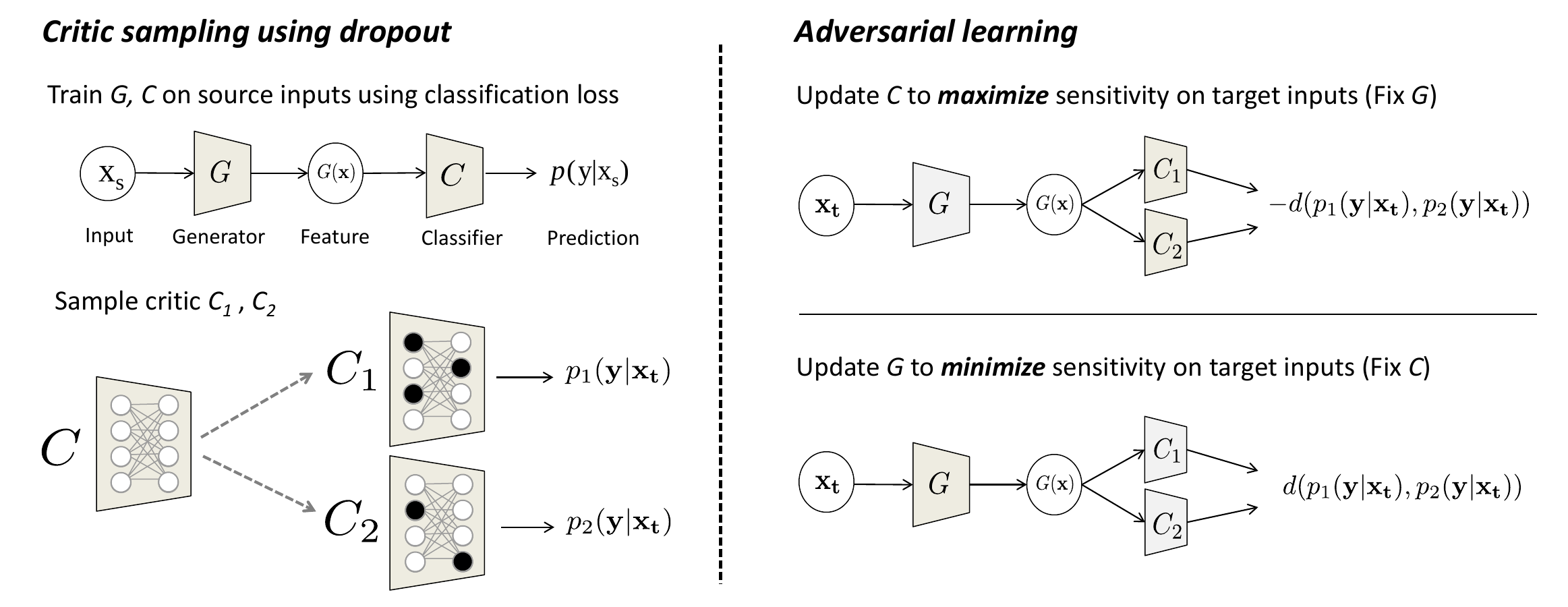}
  \end{center}
  \vspace{-3mm}
  \caption{ {\small }{\bf Left}: We train $G$, $C$ with classification loss on source and sample a critic consisting of two classifiers using dropout. The critic's sensitivity is measured as the divergence between the class predictions of $C_1$ and $C_2$ on the same input. {\bf Right}: Adversarial training iterates two steps: the critic tries to maximize the sensitivity while the generator tries to minimize it.}
  \label{fig:method_overview}
  \end{figure}

We forward input features $G(\mathbf{x_t})$ to $C$ twice, dropping different nodes each time and obtaining two different output vectors denoted as $C_1(G(\mathbf{x_t}))$, $C_2(G(\mathbf{x_t}))$. In other words, we are selecting two different classifiers $C_1$ and $C_2$ from $C$ by dropout as in Fig.~\ref{fig:method_overview}. In the figure, the corresponding posterior probabilities are indicated as $p_1(\mathbf{y}|\mathbf{x_t})$, $p_2(\mathbf{y}|\mathbf{x_t})$, abbreviated as $p_1$ and $p_2$ in the following discussion. In order to detect the change of predictions near the boundary, the critic tries to increase the difference between the predictions of $C_1$ and $C_2$. This difference corresponds to $C$'s sensitivity to the noise caused by dropout. 

To measure the sensitivity $d(p_1,p_2)$ between the two obtained  probabilistic outputs, we use the symmetric kullback leibler (KL) divergence. Formally, the divergence is calculated as
\begin{equation}
d(p_1,p_2) =  \frac{1}{2}(D_{kl}(p_1|p_2)+D_{kl}(p_2|p_1))
\label{eq:distance}
\end{equation}
where KL divergence between $p$ and $q$ is denoted as $D_{kl}(p|q)$. 

\subsection{Training Procedure}\label{sec:procedure}
\vspace{-3mm}        
In our approach, $C$ works as both critic and classifier. 
There are the following three requirements in our method:
1) $C$ and $G$ must classify source samples correctly to obtain discriminative features; 2) $C$ should maximize the sensitivity for target samples to detect the samples near the boundary; 3) $G$ should learn to minimize the sensitivity to move target samples away from the boundary.

The training within the same mini-batch consists of the following three steps.

{\bf Step 1}, in this step, $C$ is trained as a classifier. $C$ and $G$ have to classify source samples correctly to obtain discriminative features. Thus, we update both networks' parameters based on the following standard classification loss. 
%\textcolor{red}{KS: next sentence is unclear, removed} \textcolor{blue}{Kuniaki: yes, we should delete the sentence.}
%Although we use dropout in this phase, we abbreviate the notation for simplicity. 
Given source labels $y_{s}$ and samples $\mathbf{x_{s}}$, the objective in this step is
\newcommand{\mymin}{\mathop{\rm min}\limits}
\newcommand{\mymax}{\mathop{\rm max}\limits}
\newcommand{\1}{\mbox{1}\hspace{-0.25em}\mbox{l}}
\begin{equation}
   \mymin_{G,C} L(X_{s},Y_{s}) = -{\mathbb{E}_{(\mathbf{x_{s}},y_{s})\sim(X_{s},Y_{s})}}\sum_{k=1}^{K}{\1_{[k=y_{s}]}}\log p(y|{\mathbf x_{s}})%C(G(\mathbf{x_{s}}))
   \label{eq:crossentropy}
  \end{equation}
{\bf Step 2}, in this step, $C$ is trained as a critic to detect target samples near the boundary. Two classifiers are sampled from $C$ for each target sample using dropout twice to obtain $p_1$ and $p_2$. Then, $C$'s parameters are updated to maximize the sensitivity as measured by Eq.~\ref{eq:distance}. Since $C$ should learn discriminative features for source samples, in addition to the sensitivity term, we add  Eq.~\ref{eq:crossentropy}. We experimentally confirmed that this term is essential to obtain good performance.
\begin{equation}
  \mymin_{C} L(X_{s},Y_{s}) - L_{adv}(X_{t}) \\
\end{equation}
\begin{equation}
  L_{adv}(X_{t}) = {\mathbb{E}_{\mathbf{x_{t}}\sim X_{t}}}[d(p_1,p_2)]
  \label{eq:sensitivity}
  \end{equation}
{\bf Step 3}, in order to obtain representations where target samples are placed far from the decision boundary, $G$ is trained to minimize sensitivity. Here we do not add the categorical loss for source samples as in Step 2, as the generator is able to obtain discriminative features without it. We also assume that target samples should be uniformly distributed among the possible $K$ classes and add a corresponding conditional entropy term to the generator objective. 
\begin{equation}
  \mymin_{G} L_{adv}(X_{t}) + {\mathbb{E}_{\mathbf{x_{t}}\sim X_{t}}}\sum_{k=1}^{K}p(y=k|{\mathbf x_{t}}) \log p(y|{\mathbf x_{t}})\\
  \end{equation}
We update the parameters of $C$ and $G$ in every step following the defined objectives. We experimentally found it beneficial to repeat Step 3 $n$ times.
%we fixed $n=4$ in our experiments on domain adaptation.\textcolor{red}{what about GAN experiments?}
%\textcolor{blue}{Kuniaki: We set $n=1$ in GAN experiment. I mentioned it in the section of experiments on GAN.}
% KS: ok I moved this sentence to DA experiments.

\subsection{Insight and Improvement}\label{insight}
\vspace{-3mm}        
Our ADR approach encourages different neurons of the classifier  to learn different characteristics of the input (see Sec.~\ref{sec:toy}.) 
The output is the combination of shared and unshared nodes, therefore, to maximize the sensitivity, the unshared nodes must learn different features of target samples. As learning proceeds, each neuron in $C$ will capture different characteristics. At the same time, to minimize the sensitivity, $G$ learns to extract pure categorical information. If $G$ outputs features which are not related to categorical information, such as texture, slight contrast or difference of color, $C$ will utilize them to maximize sensitivity.

The trained classifier will be sensitive to the perturbation of targets caused by dropout. We note that our approach is contrary to methods called adversarial example training~(\cite{goodfellow2014explaining,miyato2015distributional}) which train the classifier to be robust to adversarial examples. They utilize input noise which can deceive or change the output of the classifier, and incorporate it to obtain a good classifier. Our ADR method encourages feature generator to obtain noise-robust target features. However, with regard to the classifier, it is trained to be sensitive to noise. To improve the final accuracy, we learn another classifier $C'$ that is not trained to be sensitive to the noise. $C'$ takes features generated by $G$ and is trained with classification loss on source samples. The loss of $C'$ is not used to update $G$. 
%$C'$ is not trained to be sensitive to the noise, thus $C'$ should be a better classifier compared to $C$. 
We compare the accuracy of $C$ and $C'$ in experiments on image classification.

\section{Experiments}
% \vspace{-3mm}        
% We perform experiments on toy datasets, unsupervised domain adaptation datasets, semi-supervised learning and image training of GAN.
\subsection{Experiment on Toy data}\label{sec:toy}
\vspace{-3mm}        
\noindent {\bf Experimental Setting.} In this experiment, we observe the decision boundary obtained by each neuron to demonstrate that ADR encourages the neurons to learn different input characteristics.
We use synthetic ``two moons'' data for this problem. Two dimensional samples from two classes are generated as source samples. Target samples are obtained by rotating the source samples. In our setting, the rotation was set to 30 degrees and data was generated with {\it scikit-learn}~(\cite{pedregosa2011scikit}). 
We train a six-layered fully-connected network; the lower 3 layers are used as feature generator, and upper 3 layers are used as classifier. We used Batch Normalization (\cite{ioffe2015batch}) and ReLU as activation function. The number of neurons are [2,5,5] for feature generator, [5,5,2] for classifier. We visualize the boundary obtained from each neuron in last layer by removing the output of all other neurons. 

\begin{figure}[t]
 \begin{minipage}{0.16\hsize}
  \begin{center}
   \includegraphics[width=\hsize]{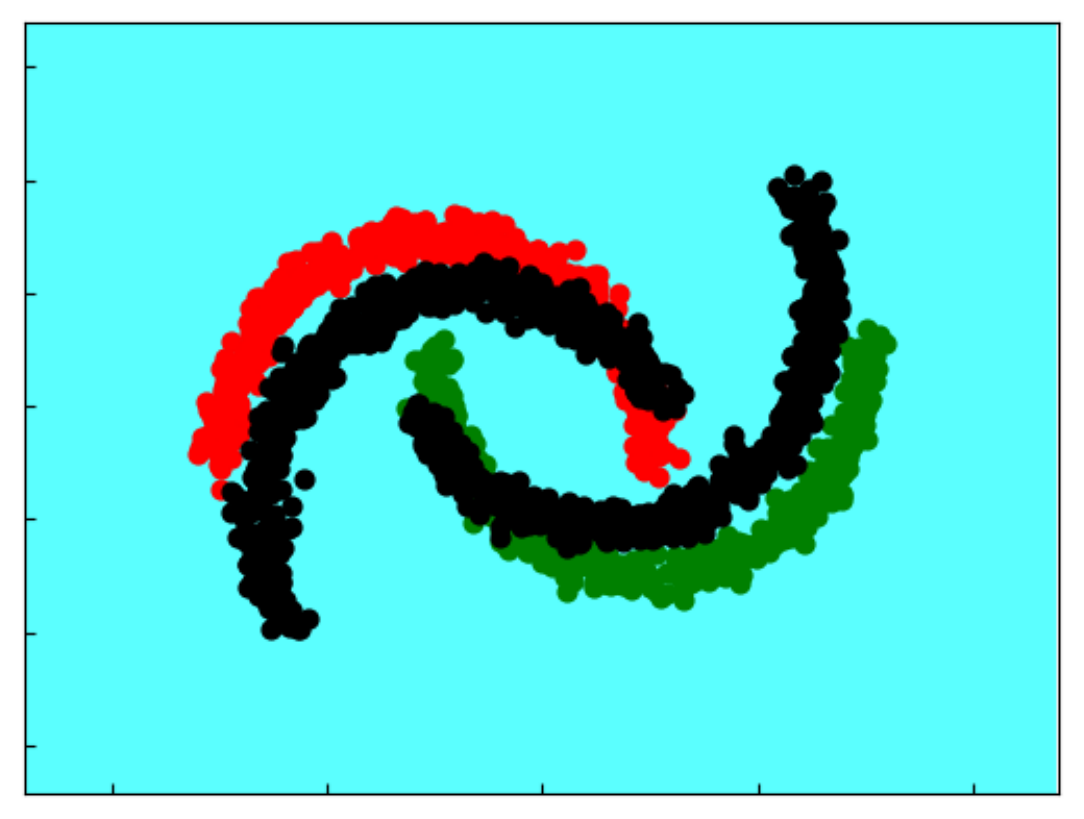}
  
  \end{center}
 \end{minipage}
 \begin{minipage}{0.16\hsize}
 \begin{center}
  \includegraphics[width=\hsize]{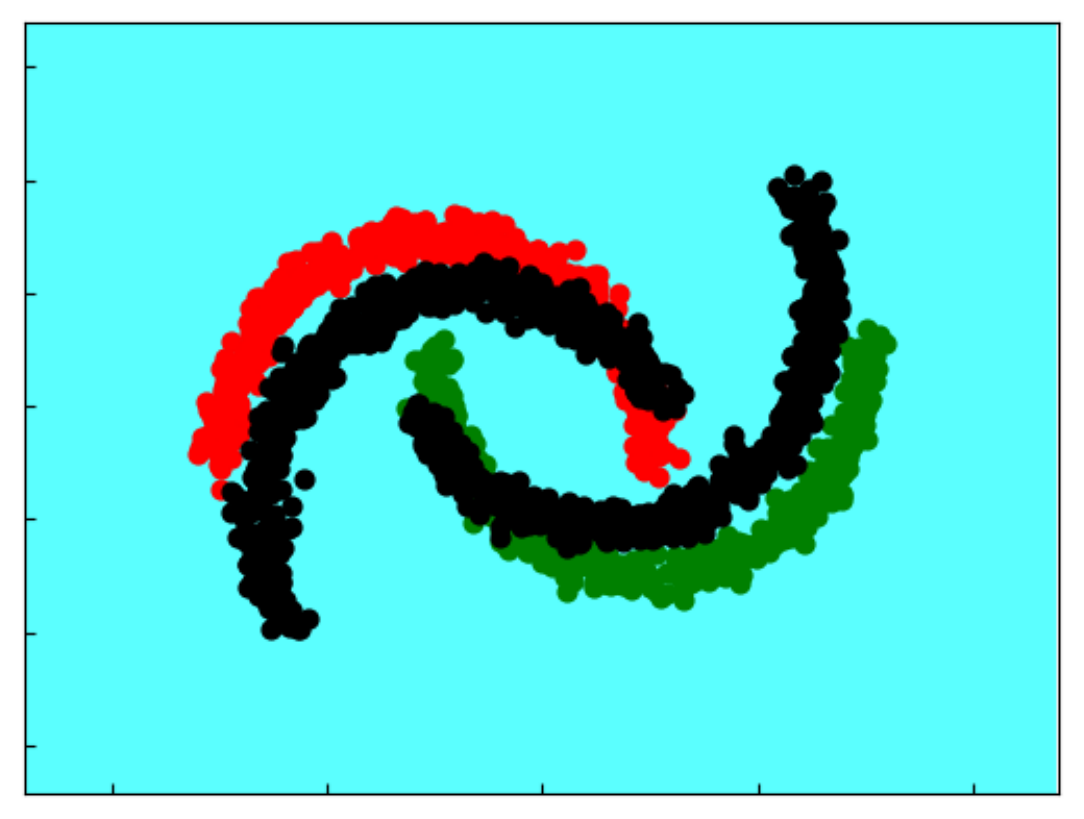}
 \end{center}
 \end{minipage}
 \begin{minipage}{0.16\hsize}
 \begin{center}
  \includegraphics[width=\hsize]{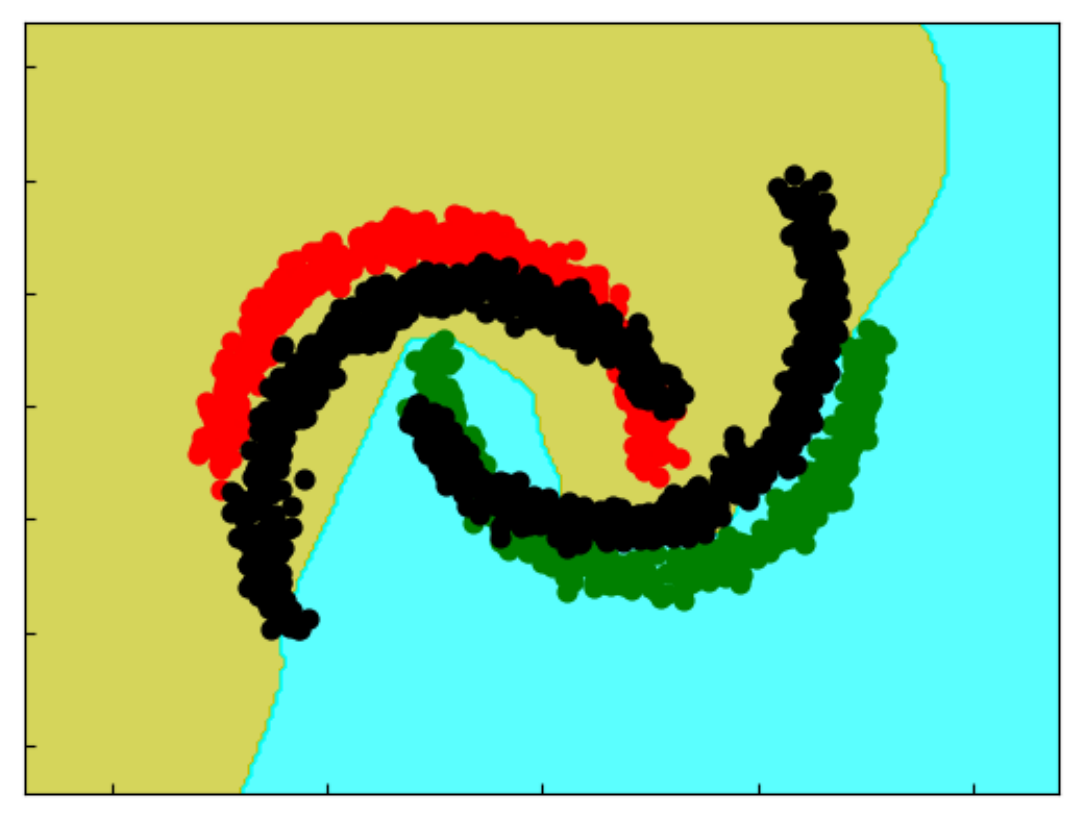}
 \end{center}
 \end{minipage}
  \begin{minipage}{0.16\hsize}
 \begin{center}
  \includegraphics[width=\hsize]{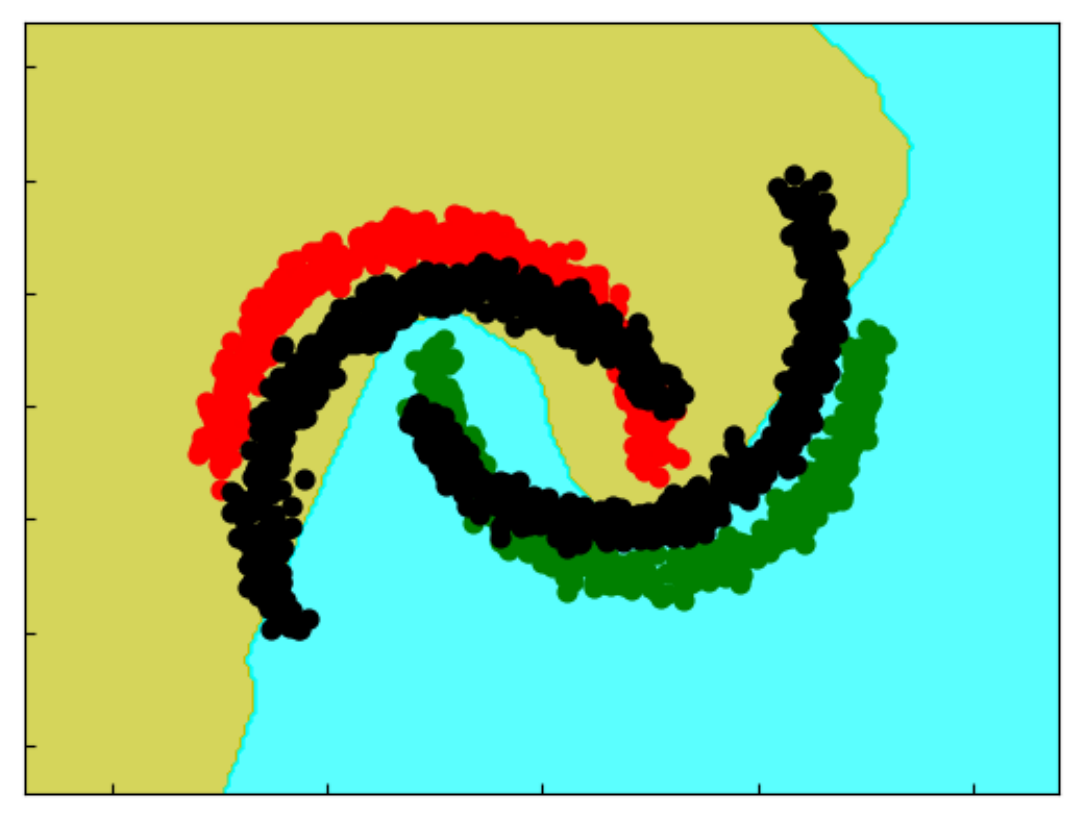}
 \end{center}
 \end{minipage}
  \begin{minipage}{0.16\hsize}
 \begin{center}
  \includegraphics[width=\hsize]{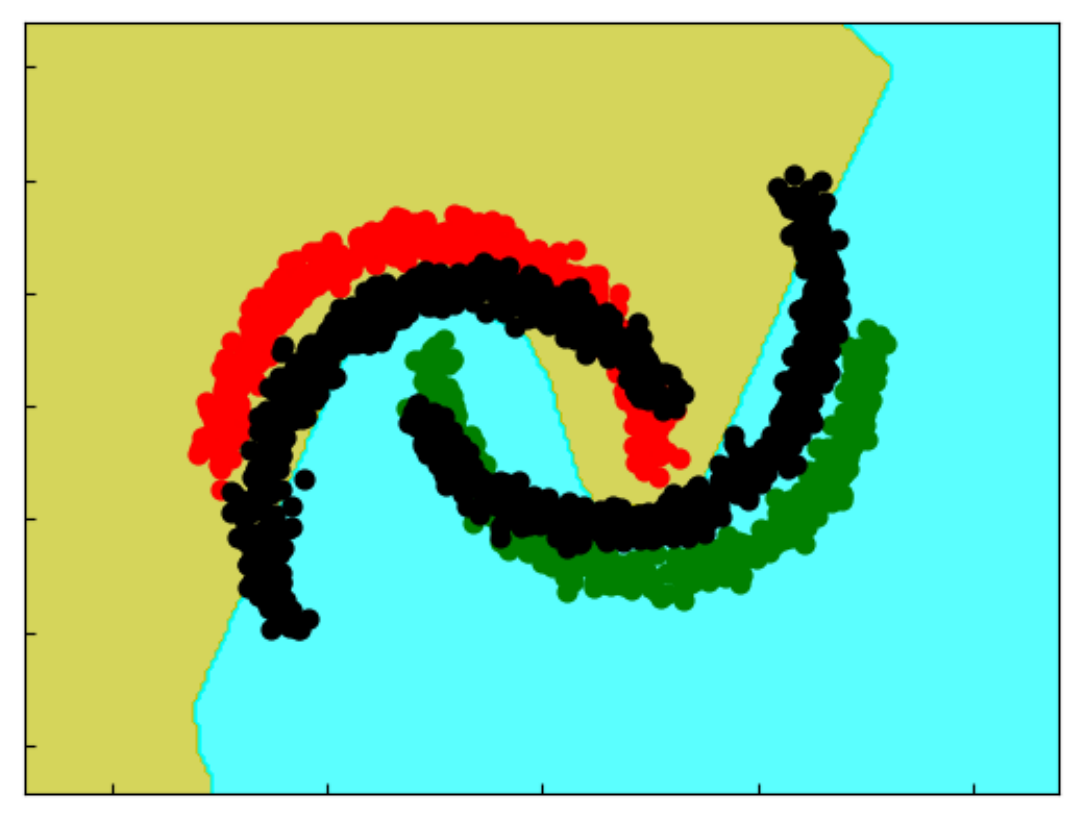}
 \end{center}
  \end{minipage}
    \begin{minipage}{0.16\hsize}
 \begin{center}
  \includegraphics[width=\hsize]{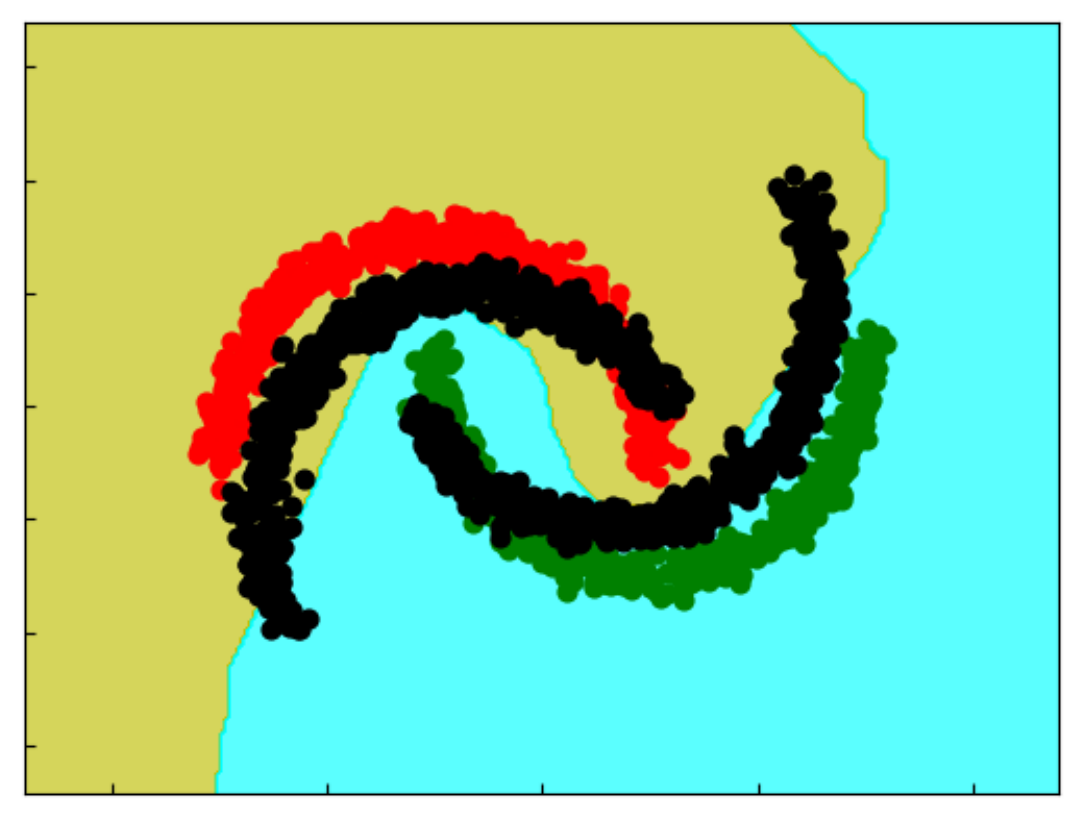}
 \end{center}
 \end{minipage}
 \begin{minipage}{0.16\hsize}
  \begin{center}
   \includegraphics[width=\hsize]{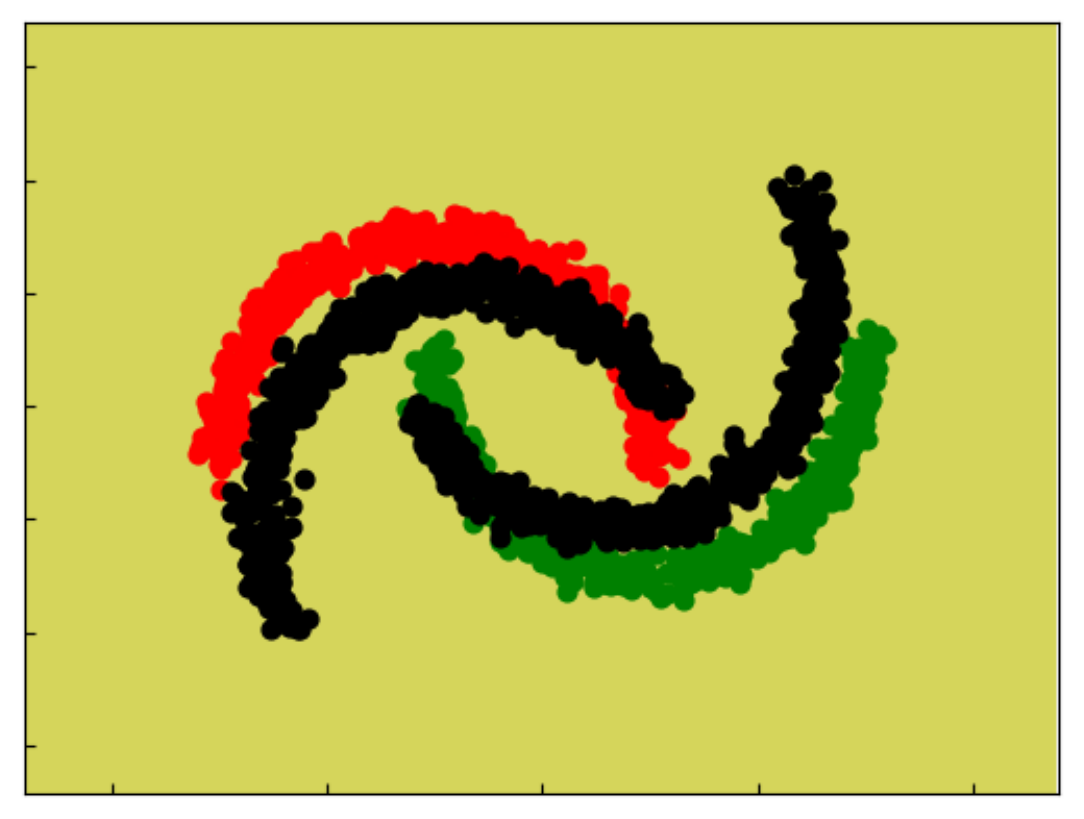}
  \end{center}
 \end{minipage}
 \begin{minipage}{0.16\hsize}
 \begin{center}
  \includegraphics[width=\hsize]{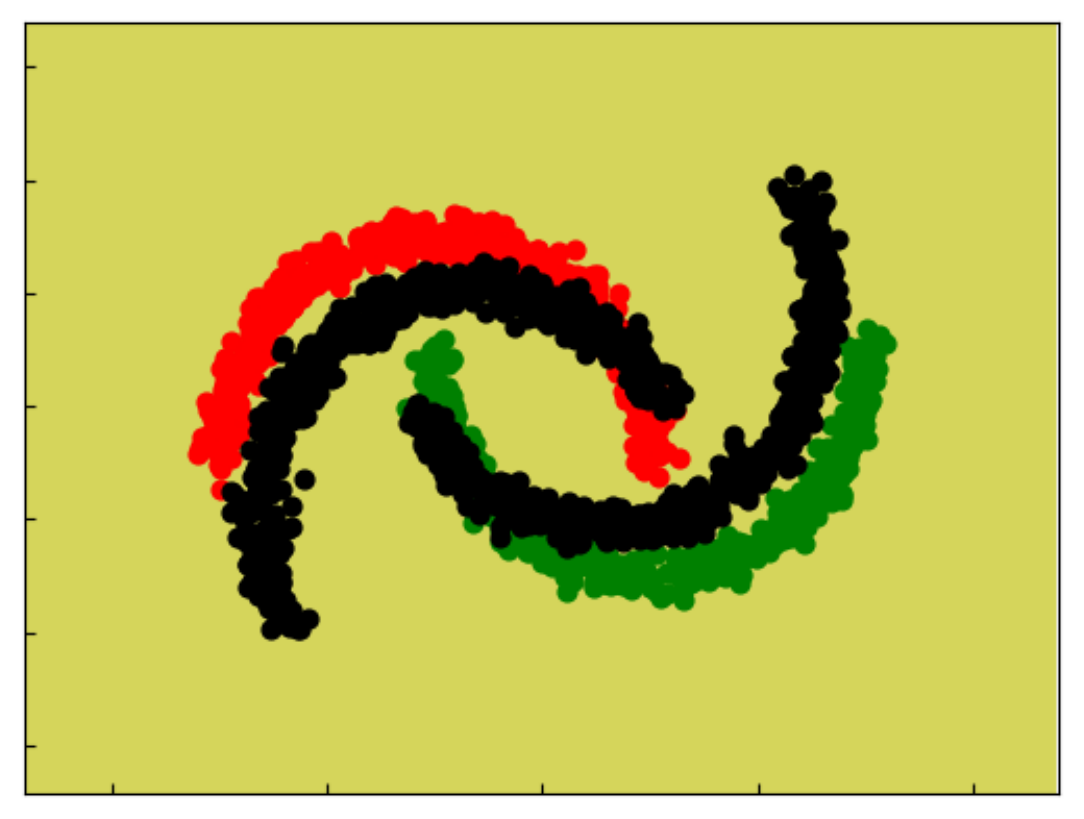}
 \end{center}
 \end{minipage}
 \begin{minipage}{0.16\hsize}
 \begin{center}
  \includegraphics[width=\hsize]{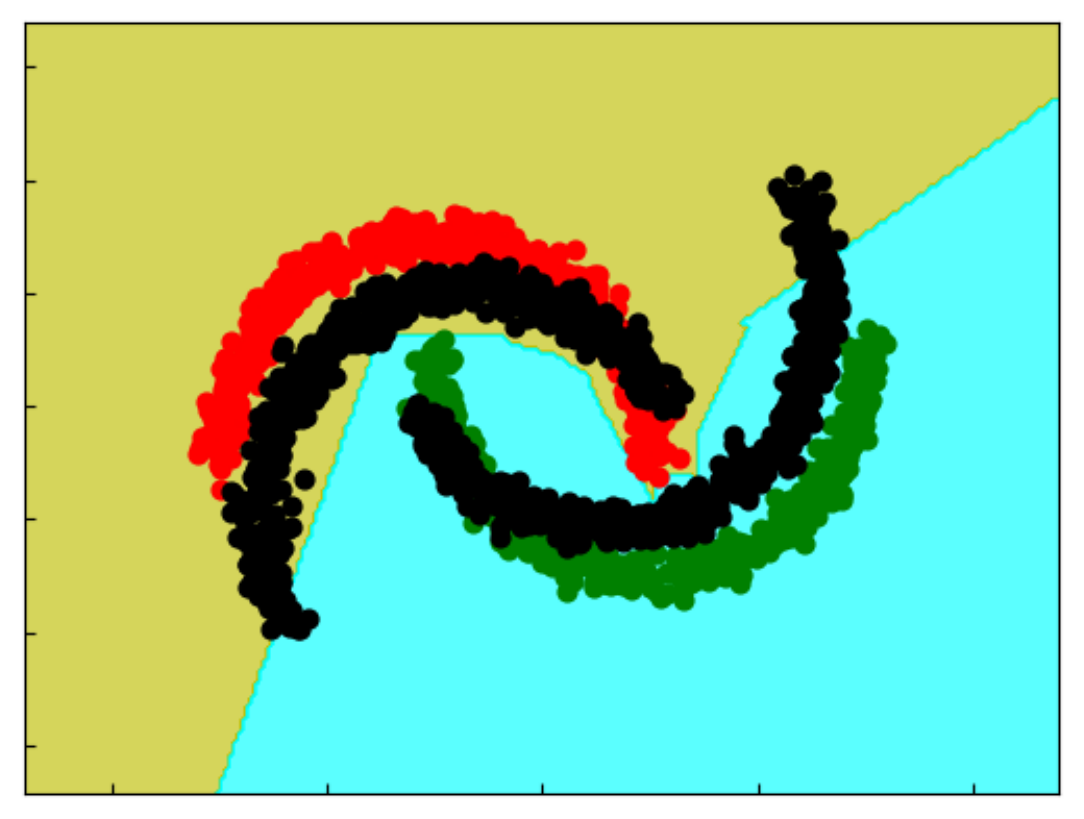}
 \end{center}
 \end{minipage}
  \begin{minipage}{0.16\hsize}
 \begin{center}
  \includegraphics[width=\hsize]{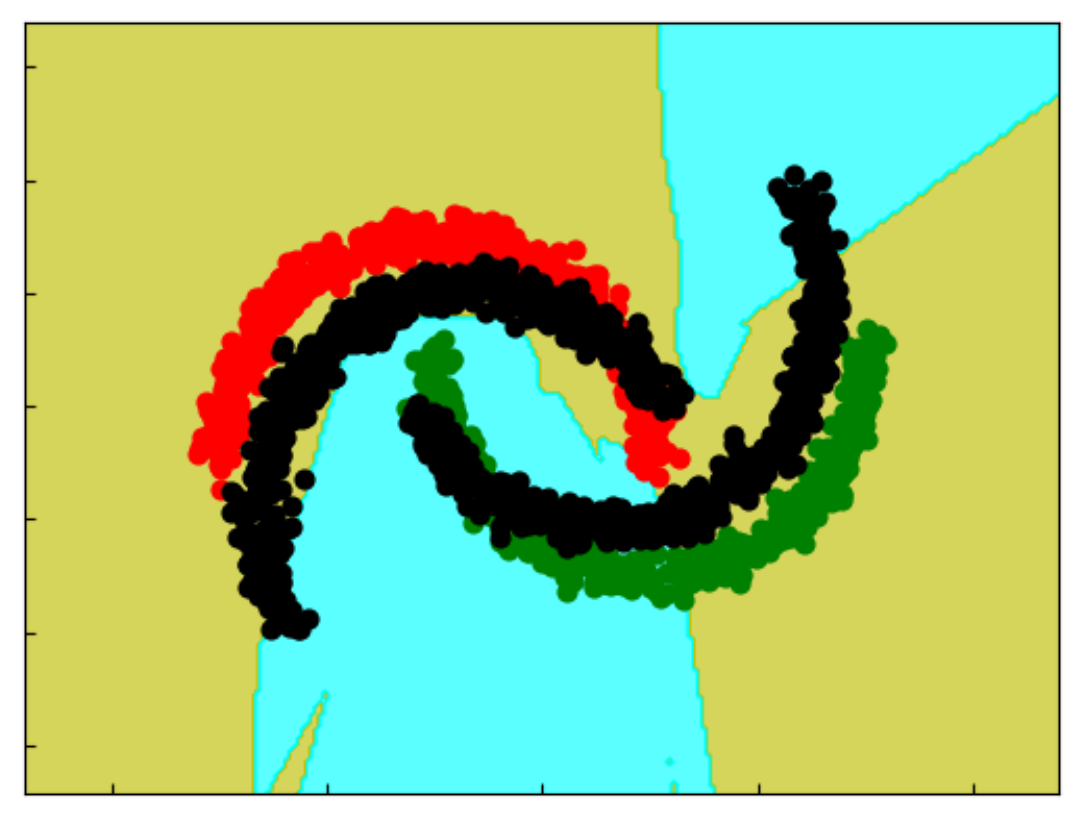}
 \end{center}
 \end{minipage}
  \begin{minipage}{0.16\hsize}
 \begin{center}
  \includegraphics[width=\hsize]{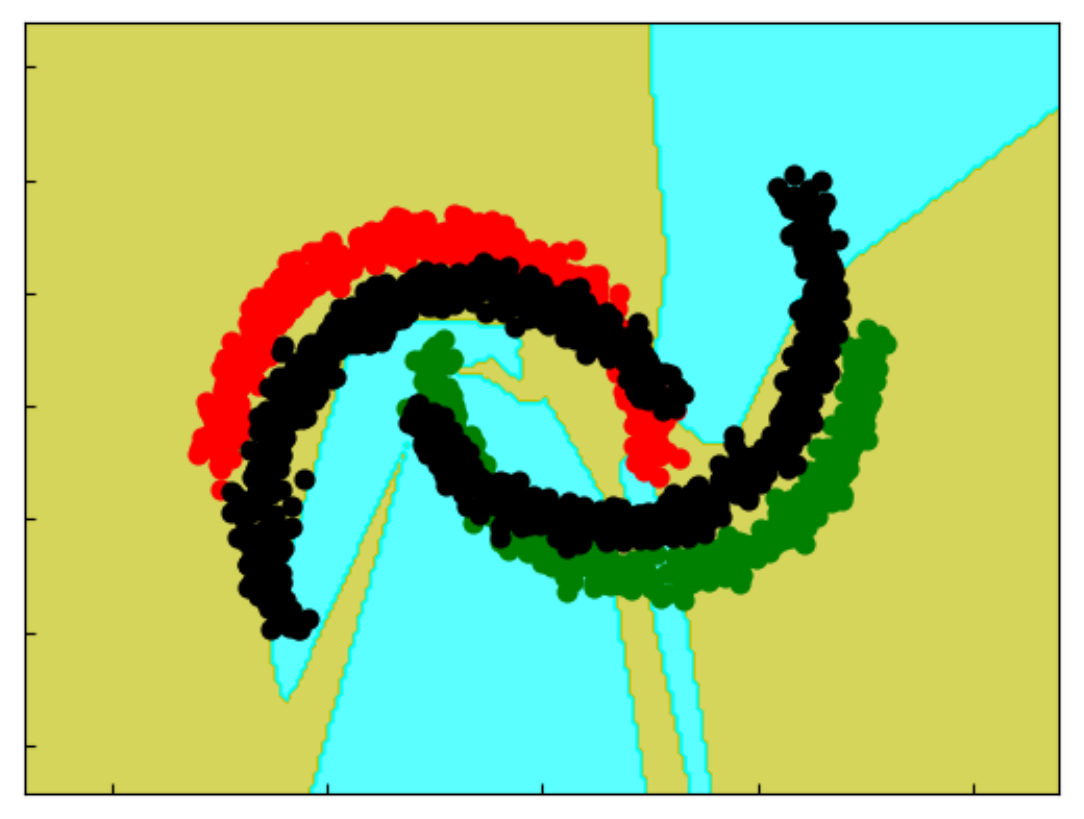}
 \end{center}
  \end{minipage}
    \begin{minipage}{0.16\hsize}
 \begin{center}
  \includegraphics[width=\hsize]{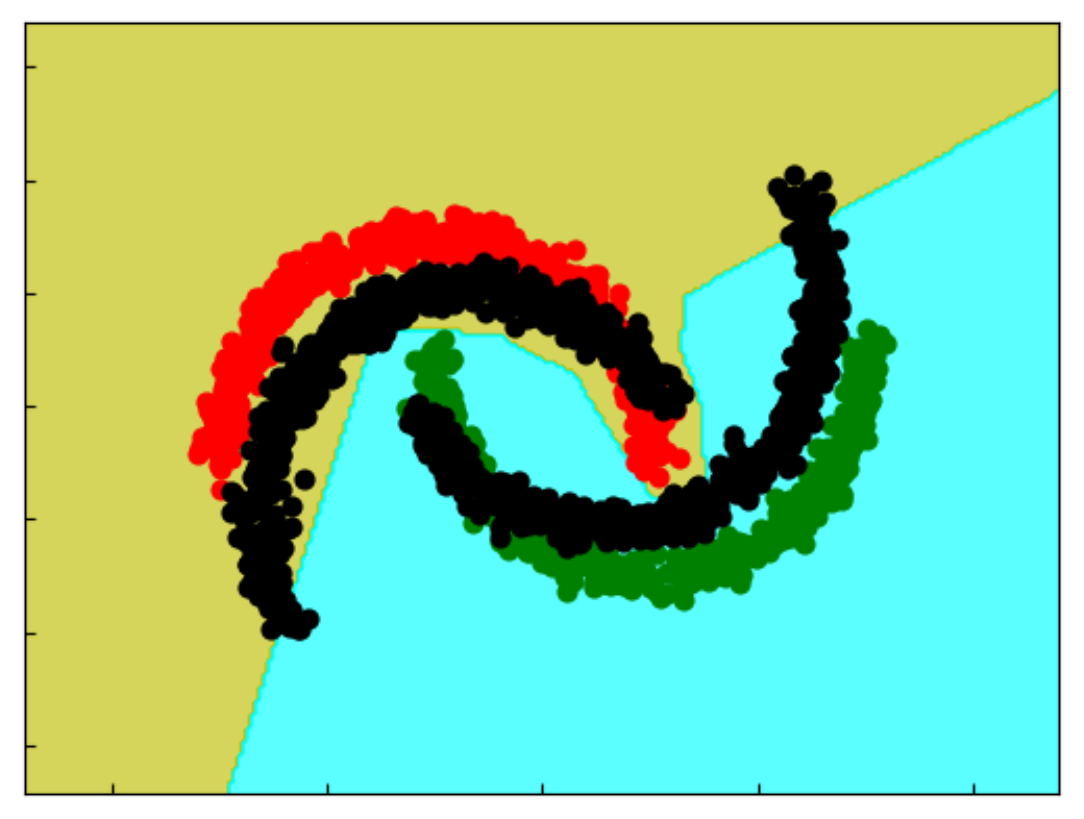}
 \end{center}
 \end{minipage}

    \caption{{\small (Best viewed in color) Toy Experiment. {\bf Top row}: Model trained without adaptation. Columns 1-5 show the decision boundary obtained by keeping one neuron in the last hidden layer and removing the rest. Red points are source samples of class one, green points are class two. Black points are target samples. The yellow region indicates where the samples are classified as class one, cyan region class two. We see that the neurons do not learn very diverse features. Column 6 shows the boundary obtained by keeping all 5 neurons. 
    {\bf Bottom row}: Boundaries learned by the model adapted by our adversarial dropout method. Unlike the top row, here neurons 3,4,5 learn diverse features which result in diverse boundaries. } }
\label{fig:toy_data}
\vspace{-3mm}
\end{figure}

\noindent {\bf Results.} We show the learned boundary in Fig.~\ref{fig:toy_data}. In the baseline model trained only with source samples (top row), two of five neurons do not seem to learn an effective boundary, and three neurons learn a similar boundary. On the other hand, in our method (bottom row), although two neurons do not seem to learn any meaningful boundary, three neurons learn a distinctive boundaries. Each neuron is trained to be sensitive to the noise caused by target samples. The final decision boundary (rightmost column) classifies most target samples correctly. The accuracy of our proposed method is 96\% whereas the accuracy of the non-adapted model was 84\%. 

\subsection{Unsupervised Domain Adaptation for Classification}

\noindent {\bf Experiments on Digits Classification.}
We evaluate our model on adaptation between digits datasets. We use MNIST~(\cite{lecun1998gradient}), SVHN~(\cite{netzer2011reading}) and USPS datasets and follow the protocol of unsupervised domain adaptation used by~(\cite{tzeng2017adversarial}). We assume no labeled target samples and use fixed hyper-parameters for all experiments, unlike other works that use a target validation set~(\cite{saito2017asymmetric}). The number of iterations for Step 3 was fixed at $n=4$.
We used the same network architecture as in (\cite{tzeng2017adversarial}), but inserted a Batch Normalization layer before the activation layer to stabilize the training. We used Adam~(\cite{kingma2014adam}) for optimizer and set the learning rate to $2.0\times10^{-4}$. We set the learning rate to a value commonly reported in the GAN literature.
%\textcolor{blue}{We set the value because other papers dealing with GAN often this learning rate. Although we did not try other learning rate exhaustively, we think the performance of our method will not decrease due to the change in the learning rate. The performance may become even better.}

We compare our approach to several existing methods and to the entropy minimization baseline (ENT) obtained by modifying~(\cite{springenberg2015unsupervised}). Due to space limitations, we provide a detailed explanation of this baseline in the appendix.

\begin{figure}[t]
\begin{minipage}{0.33\hsize}
  \centering
  \subfigure[USPS to MNIST]{\includegraphics[width=\hsize]{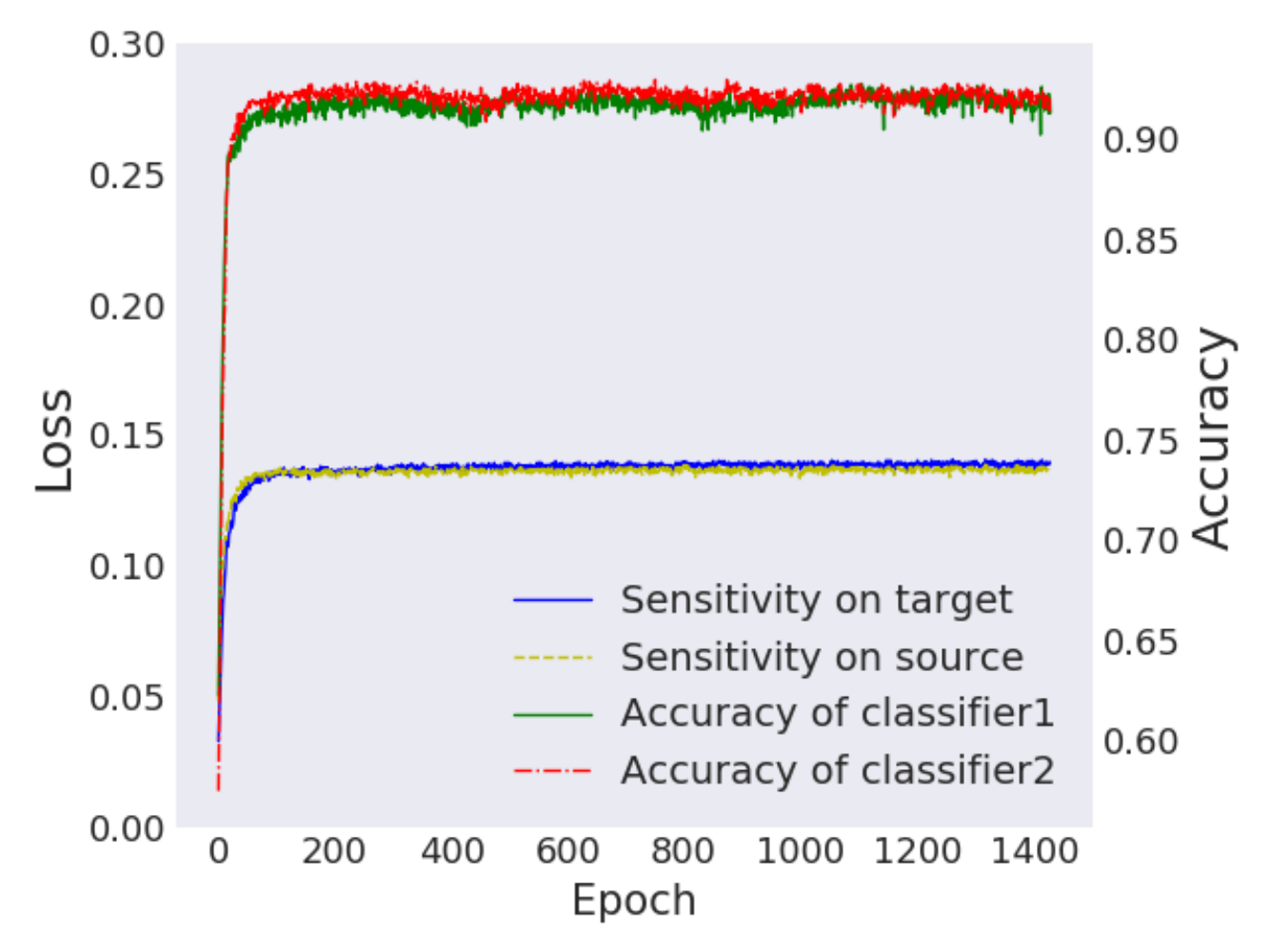}\label{fig:mnist2usps_ent}}

\end{minipage}
\begin{minipage}{0.33\hsize}
  \centering
  \subfigure[MNIST to USPS]{\includegraphics[width=\hsize]{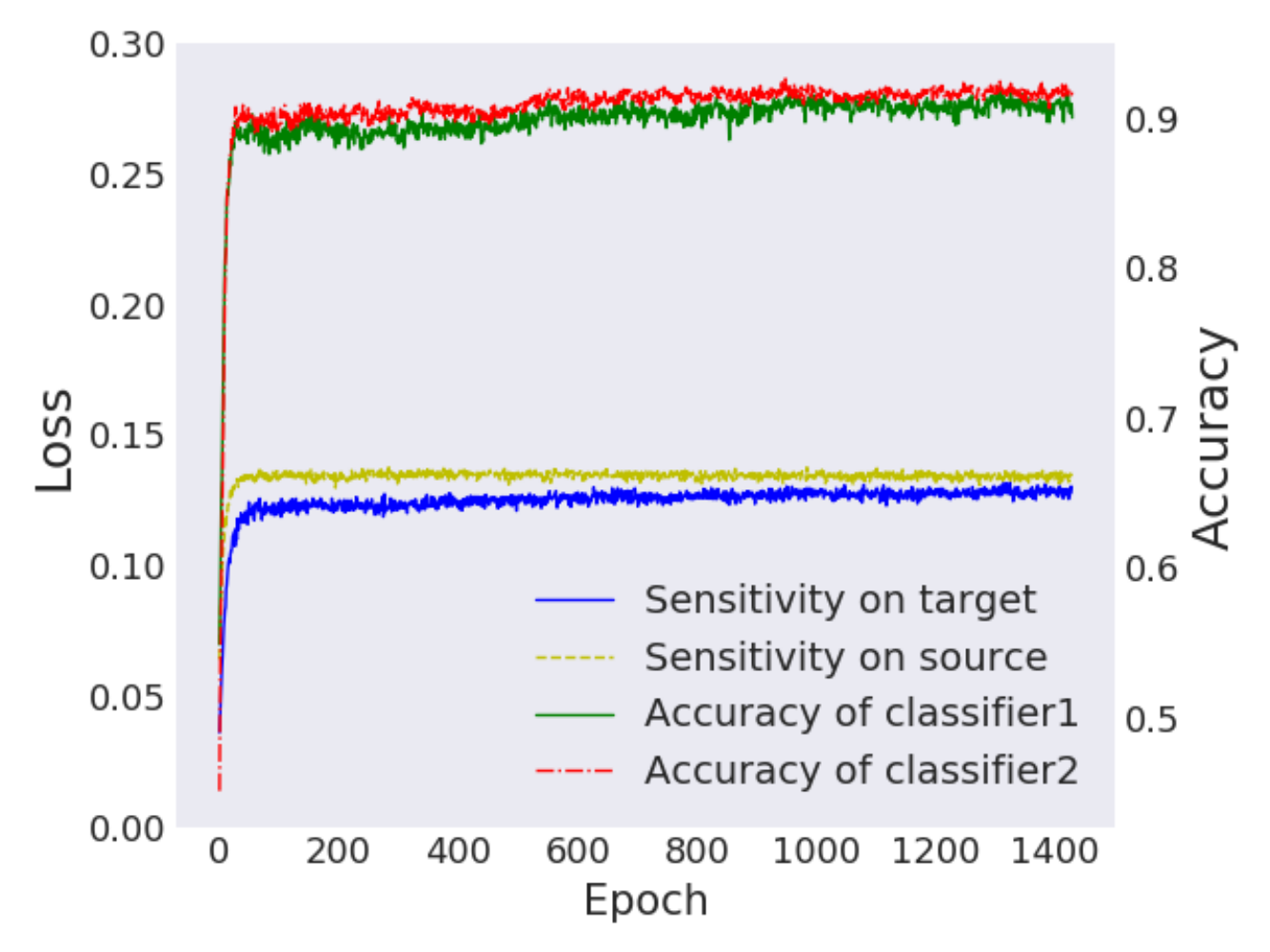}\label{fig:usps2mnist_ent}}
\end{minipage}
\begin{minipage}{0.33\hsize}
\centering
  \subfigure[SVHN to MNIST]{\includegraphics[width=\hsize]{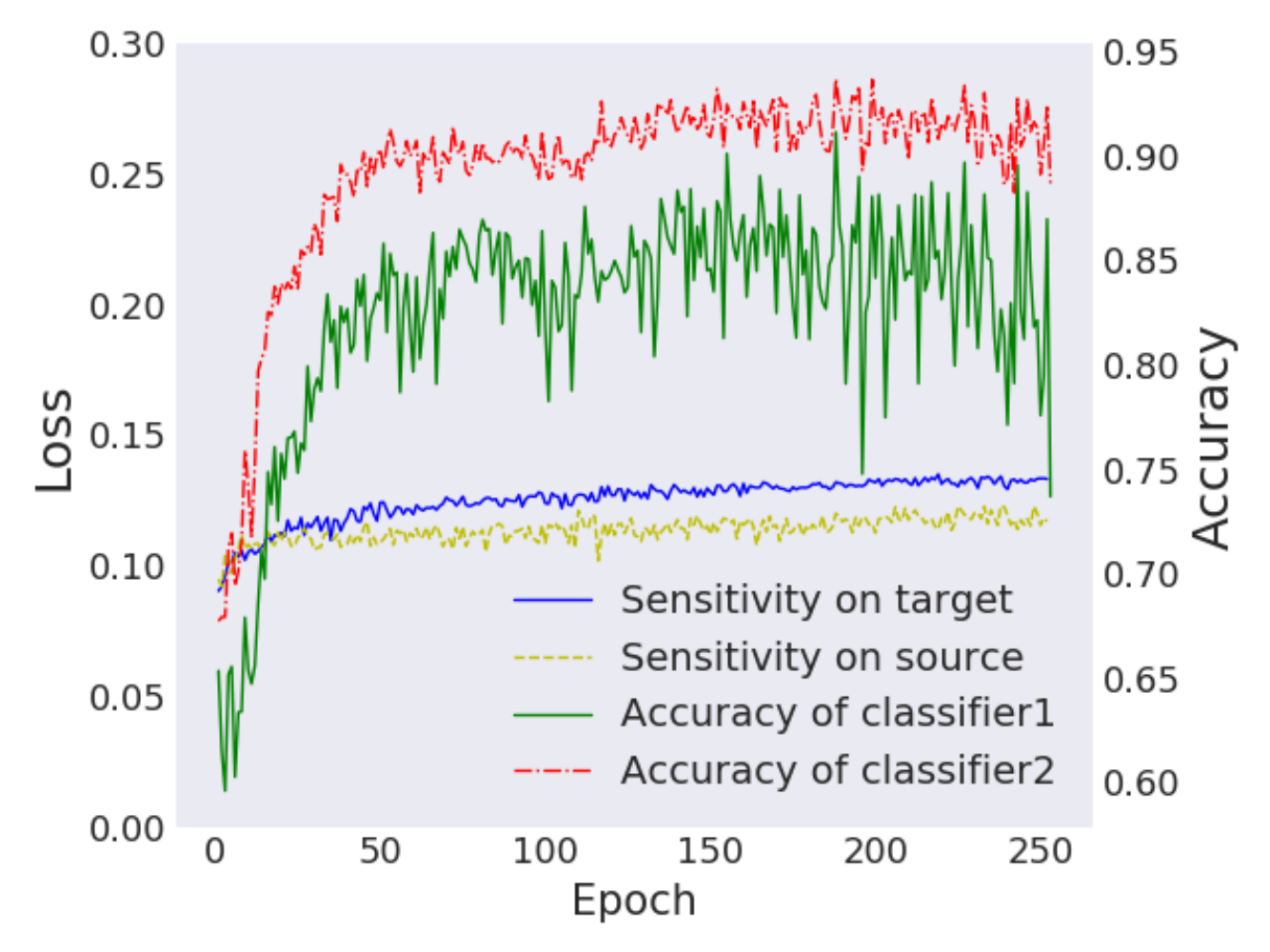}  \label{fig:svhn2mnist_ent}}

\end{minipage}
   \vspace{-5mm}
\caption{{\small Relationship between sensitivity loss on target ({\bf blue} line), on source ({\bf yellow} line), and accuracy ({\bf red}: accuracy of $C{'}$, {\bf green}: accuracy of $C$) during training on digits.}}
\label{fig:entropy}
\vspace{-4mm}
\end{figure}

\begin{figure}[t]
\begin{minipage}{0.32\hsize}
  \centering
  \subfigure[USPS to MNIST]{\includegraphics[width=\hsize]{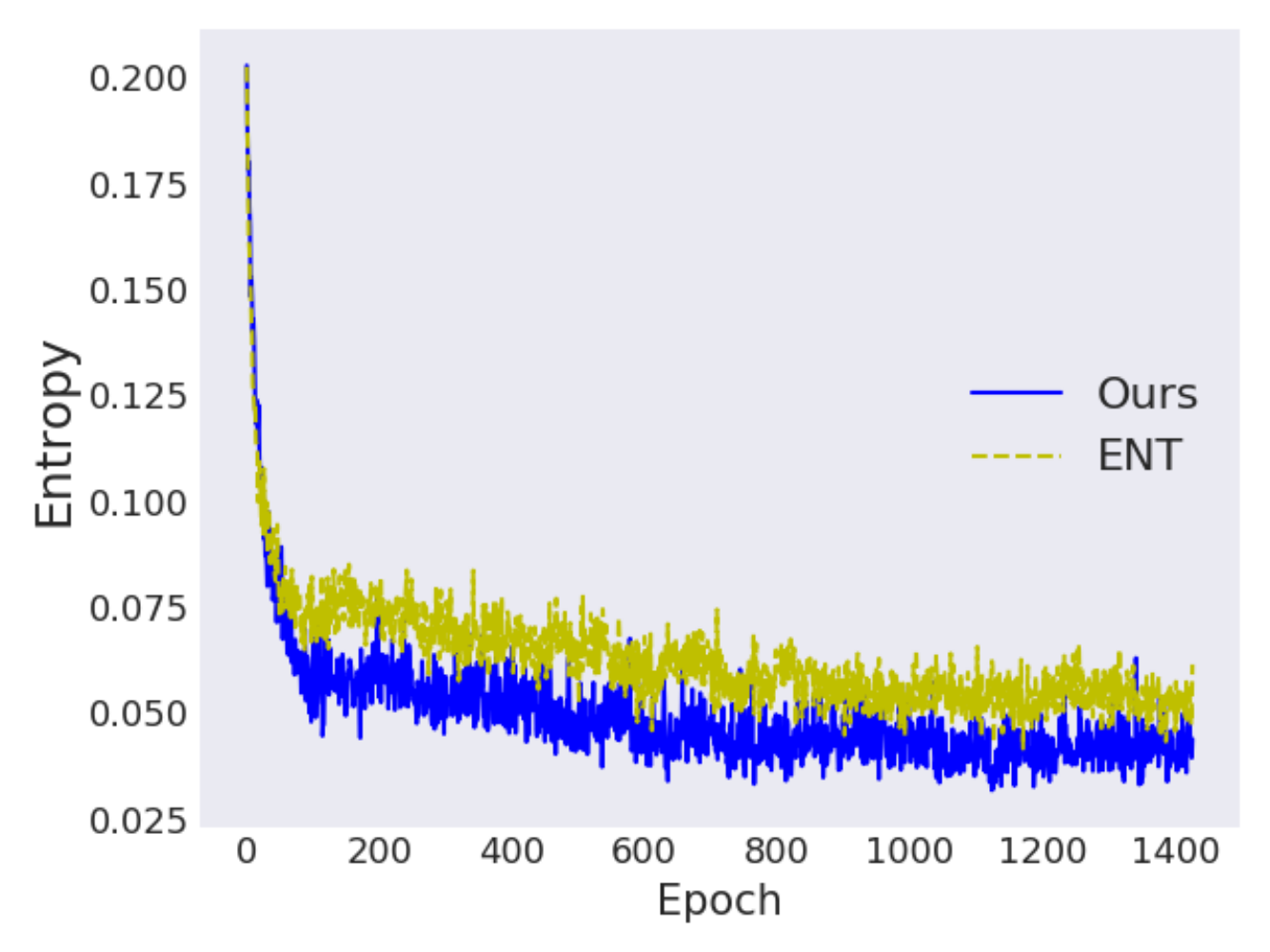}\label{fig:mnist2usps_graph}}

\end{minipage}
\begin{minipage}{0.32\hsize}
  \centering
  \subfigure[MNIST to USPS]{\includegraphics[width=\hsize]{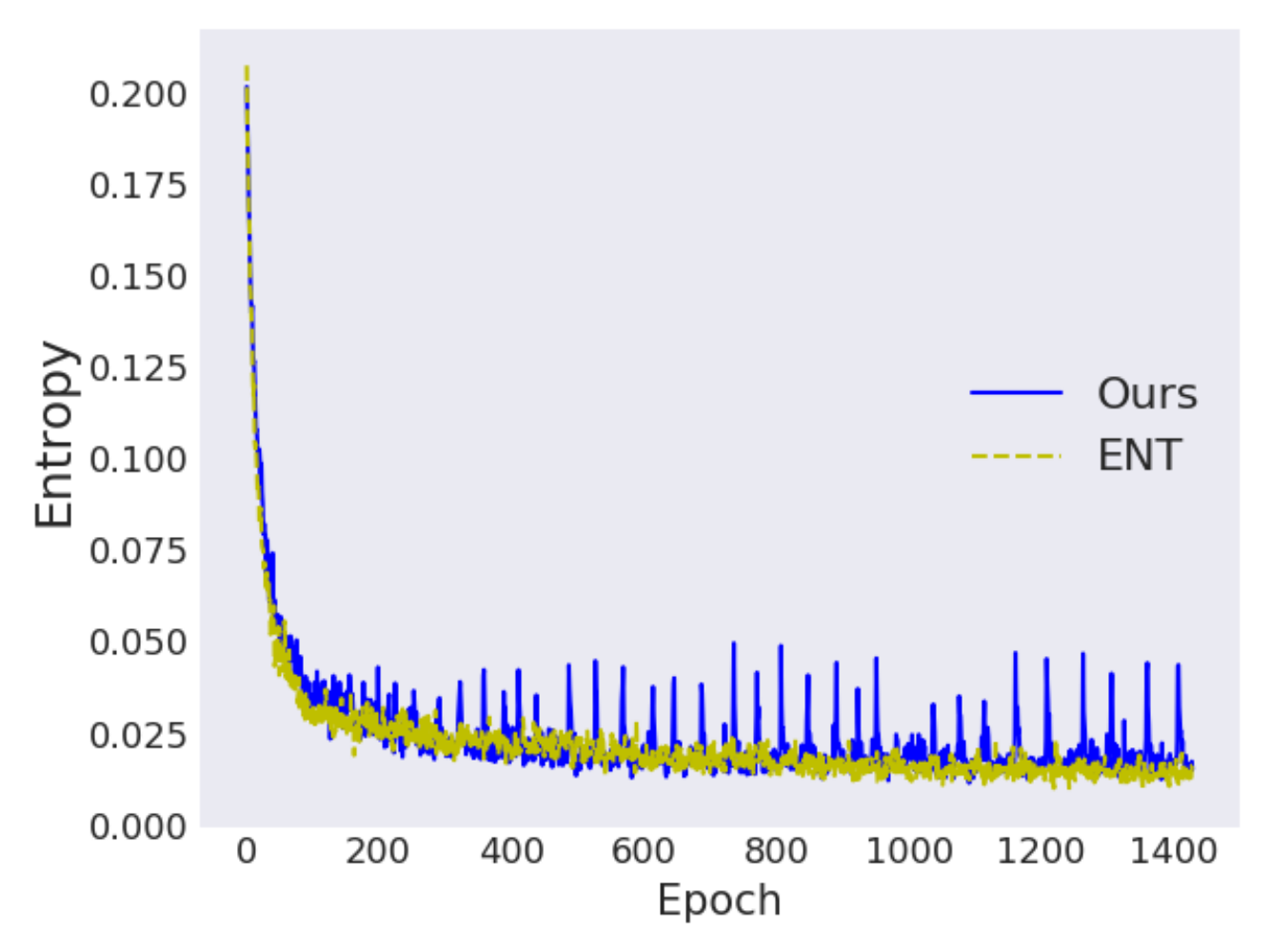}\label{fig:usps2mnist_graph}}
\end{minipage}
\begin{minipage}{0.32\hsize}
\centering
  \subfigure[SVHN to MNIST]{\includegraphics[width=\hsize]{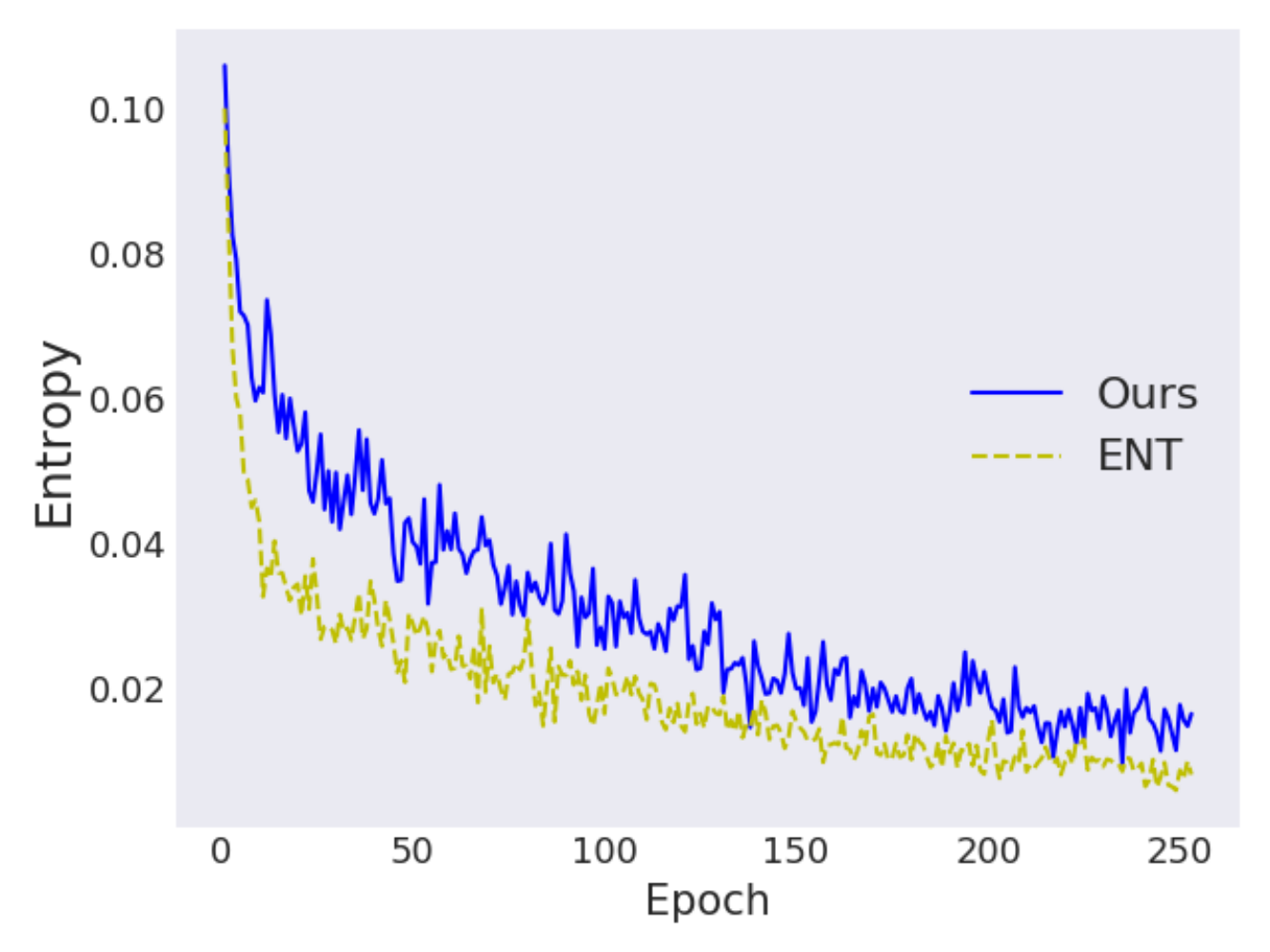}  \label{fig:svhn2mnist_graph}}
\end{minipage}
   \vspace{-4mm}
\caption{{\small Comparision of entropy of ours ({\bf blue} line) with ENT ({\bf yellow} line). The entropy is calculated on target samples by using the output of the classifier.}}
\label{fig:acc_graph}
\vspace{-3mm}
\end{figure}

Results in Table~\ref{table:exp_scratch} demonstrate that ADR obtains better performance than existing methods. In particular, on the challenging adaptation task from SVHN to MNIST, our method achieves much better accuracy than previously reported. Fig. \ref{fig:acc_graph} shows the learning curve of each experiment. As sensitivity loss increases, the target accuracy improves. This means that as critic $C$ learns to detect the non-discriminative samples, feature generator $G$ learns to fool it, resulting in improved accuracy.
In addition, we can see that the sensitivity of source samples increases too. As mentioned in Sec \ref{insight}, the critic network should learn to capture features which are not very important for classification, such as texture or slight edges, and it seems to also capture such information in source samples.
The accuracy of the classifier $C'$ (denoted by red), which is trained not to be sensitive to the noise, is almost always better than the accuracy of the critic network. In adaptation from SVHN to MNIST (Fig.~\ref{fig:svhn2mnist_graph}), the accuracy of the critic often suffers as it becomes too sensitive to the noise caused by dropout. On the other hand, the accuracy shown by the red line is stable. Our ENT baseline shows good performance compared to other existing methods. This result indicates the effectiveness of methods based on entropy minimization. In Fig. \ref{fig:entropy}, we compare our proposed method and ENT in terms of entropy of target samples. Our method clearly decreases the entropy, because target samples are moved away from the decision boundary. Yet, its behavior is different from ENT. 
Interestingly, the entropy is made smaller than ENT in case of adaptation from USPS to MNIST (Fig. \ref{fig:usps2mnist_ent}) though ENT directly minimizes the entropy and our method does not. On the SVHN to MNIST task (Fig. \ref{fig:svhn2mnist_ent}), the entropy of ADR is larger than ENT, which indicates that our method places the target samples closer to the decision boundary than ENT does. 

\begin{table}[h]
  \begin{center}
    \scalebox{0.9}{
  \begin{tabular}{l|ccccccc}
\toprule
& {\bf SVHN}&{\bf USPS}&{\bf MNIST}\\
{\bf METHOD}&to&to&to\\
 & {\bf MNIST} &{\bf MNIST}&{\bf USPS}\\ \hline
Source Only &67.1&68.1&77.0&\\\hline
ATDA~(\cite{saito2017asymmetric})&86.2$\dagger$&-&-\\
% DSN~(\cite{bousmalis2016domain})&82.7&-&-&91.2\\\hline
 DANN~(\cite{ganin2014unsupervised})&73.9&73.0$\pm$2.0&77.1$\pm$1.8\\
 DoC~(\cite{tzeng2014deep})&68.1$\pm$0.3&66.5$\pm$3.3&79.1$\pm$0.5\\
 ADDA~(\cite{tzeng2017adversarial})&76.0$\pm$1.8&90.1$\pm$0.8&89.4$\pm$0.2\\\hline
  CoGAN~(\cite{liu2016coupled})& did not converge &89.1$\pm$0.8&91.2$\pm$0.8\\
  DTN~(\cite{taigman2016unsupervised})&84.7&-&-\\\hline
  ENT &93.7$\pm$2.05&89.2$\pm$3.07&89.6$\pm$1.44\\
 Ours &{\bf 94.1}$\pm$1.37&{\bf 91.5}$\pm$3.61&{\bf 91.3}$\pm$0.65\\
 \hline
  \end{tabular}}
    \caption{{\small Results on digits datasets. Please note that $\dagger$ means the result obtained using a few labeled target samples for validation. The reported accuracy of our method is obtained from $C{'}$. ENT is our proposed baseline method. We implement entropy minimization based adversarial training as we mentioned in Appendix \ref{append:ent}.} }
\label{table:exp_scratch}
\end{center}
\vspace{-3mm}
\end{table}

\noindent {\bf Experiments on Object Classification.}
We next evaluate our method on fine-tuning a pretrained CNN. We use a new domain adaptation benchmark called the VisDA Challenge~(\cite{visda}) which focuses on the challenging task of adapting from synthetic to real images. The source domain consists of 152,409 synthetic 2D images from 12 object classes rendered from 3D models. The validation and test target domains consists of real images, which belong to the same classes. We used the validation domain (55,400 images) as our target domain in an unsupervised domain adaptation setting. 

We evaluate our model on fine-tuning networks pretrained on ImageNet~(\cite{deng2009imagenet}):
 ResNet101~(\cite{he2016deep}) and ResNext~(\cite{xie2016aggregated}). For the feature generator, we use the pre-trained CNN after removing the top fully connected layer. For the classification network, we use a three-layered fully connected network.

Table~\ref{table:exp_class} shows that our method outperformed other distribution matching methods and our new baseline (ENT) in finetuning both networks by a large margin. ENT did not achieve better performance than existing methods, though improvement over the source only model was observed. Although this method performed well on digits, it does not work as well here, possibly because of the larger shift between very different domains. 
%Entropy-minimization forces the target samples to be far away from the boundary as source samples are. But, due to the large difference between domains, clearly separating target samples into each class is difficult. This is the possible reason why the new baseline does not work well in this experimental setting.
In this experiment, after training $G$ and $C$, we retrained a classifier $C^{'}$ just on the features generated by $G$ due to GPU memory limitations, and observed improvement in both networks. 

\begin{figure}[tb]
\vspace{-4mm}
\centering
\begin{minipage}{0.3\hsize}
  \centering
  \subfigure[Pretrained Model]{\includegraphics[width=\hsize]{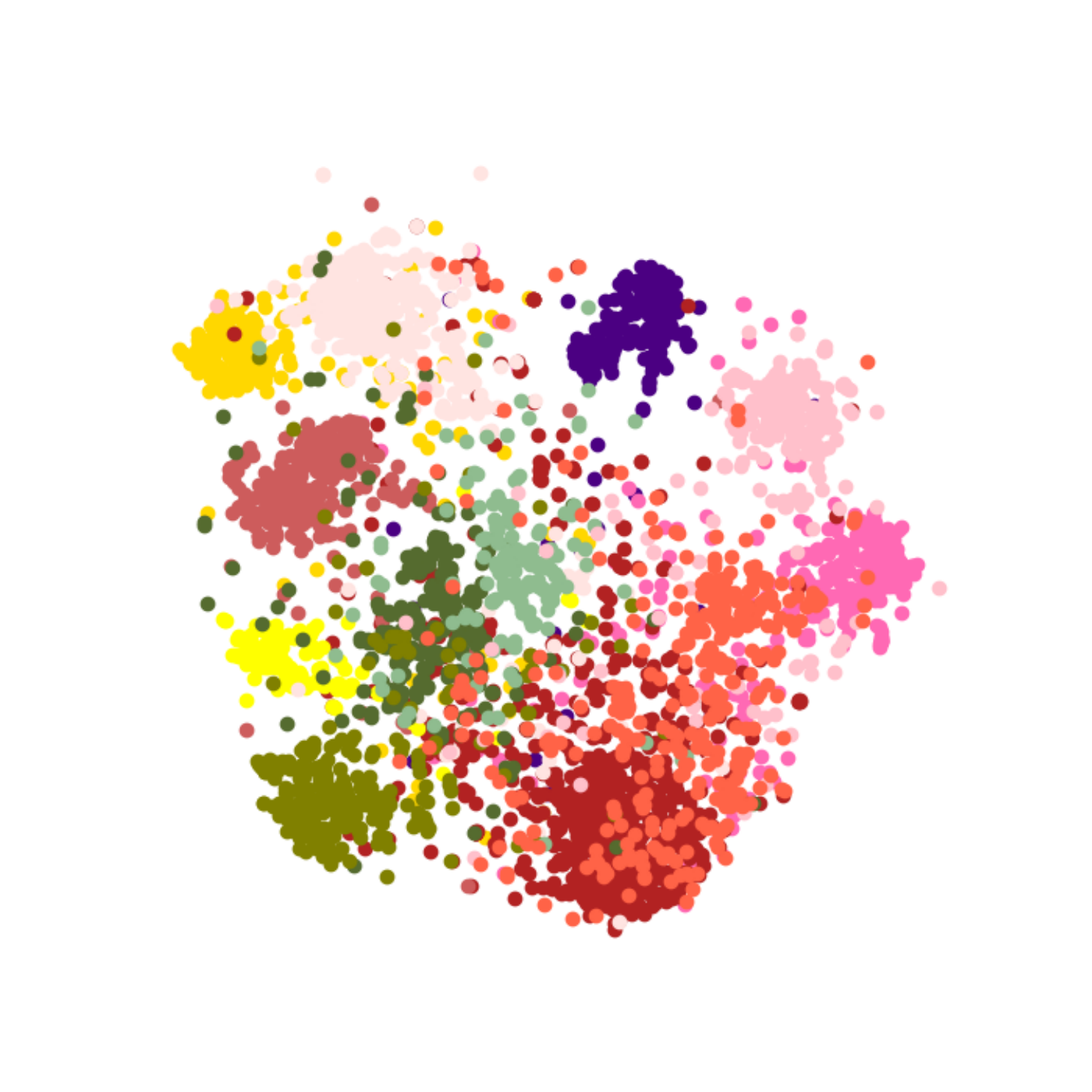}\label{fig:plot_original}}
\end{minipage}
\begin{minipage}{0.3\hsize}
  \centering
  \subfigure[Source Only Model]{\includegraphics[width=\hsize]{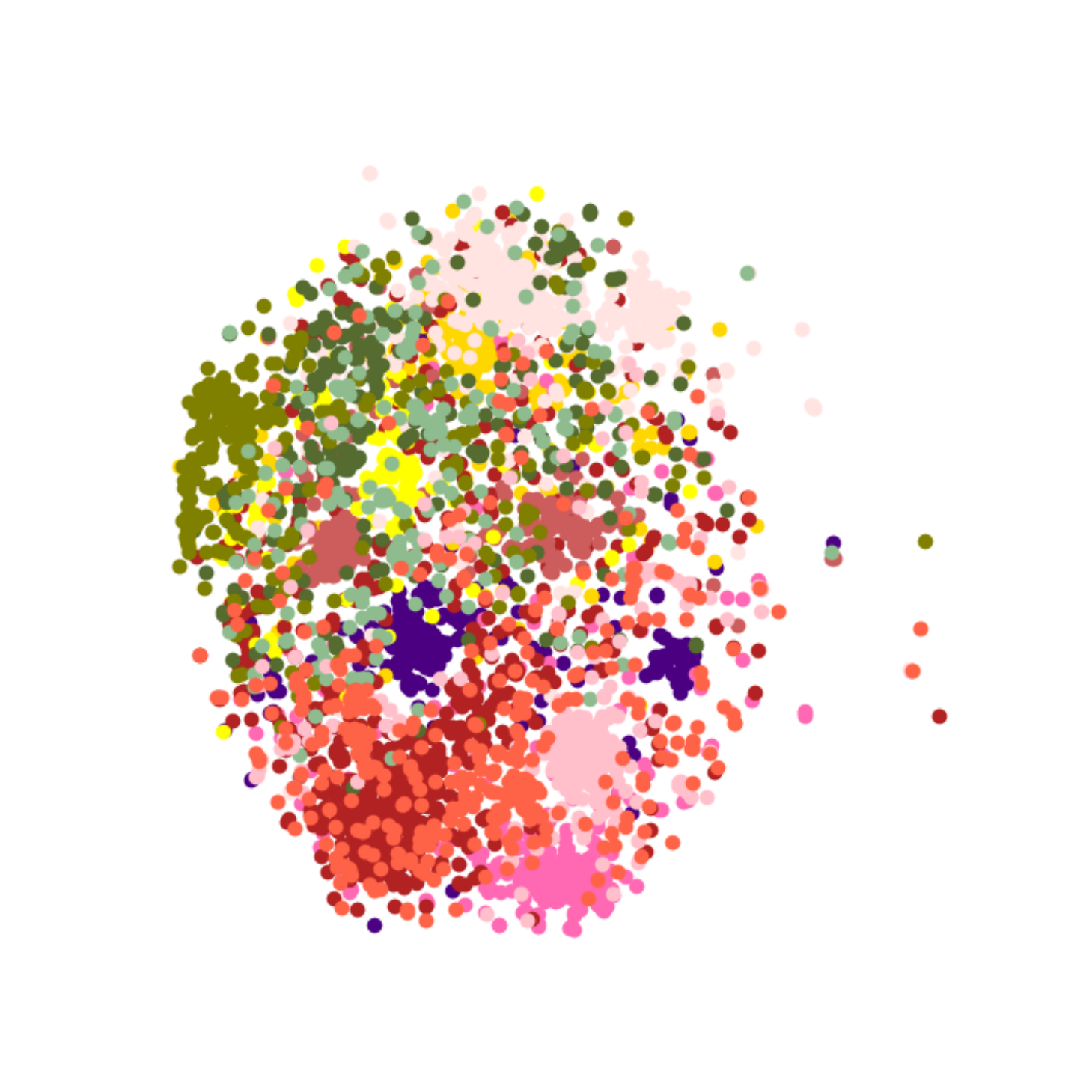}\label{fig:plot_nonadapted}}
\end{minipage}
\begin{minipage}{0.3\hsize}
  \centering
  \subfigure[Adapted Model]{\includegraphics[width=\hsize]{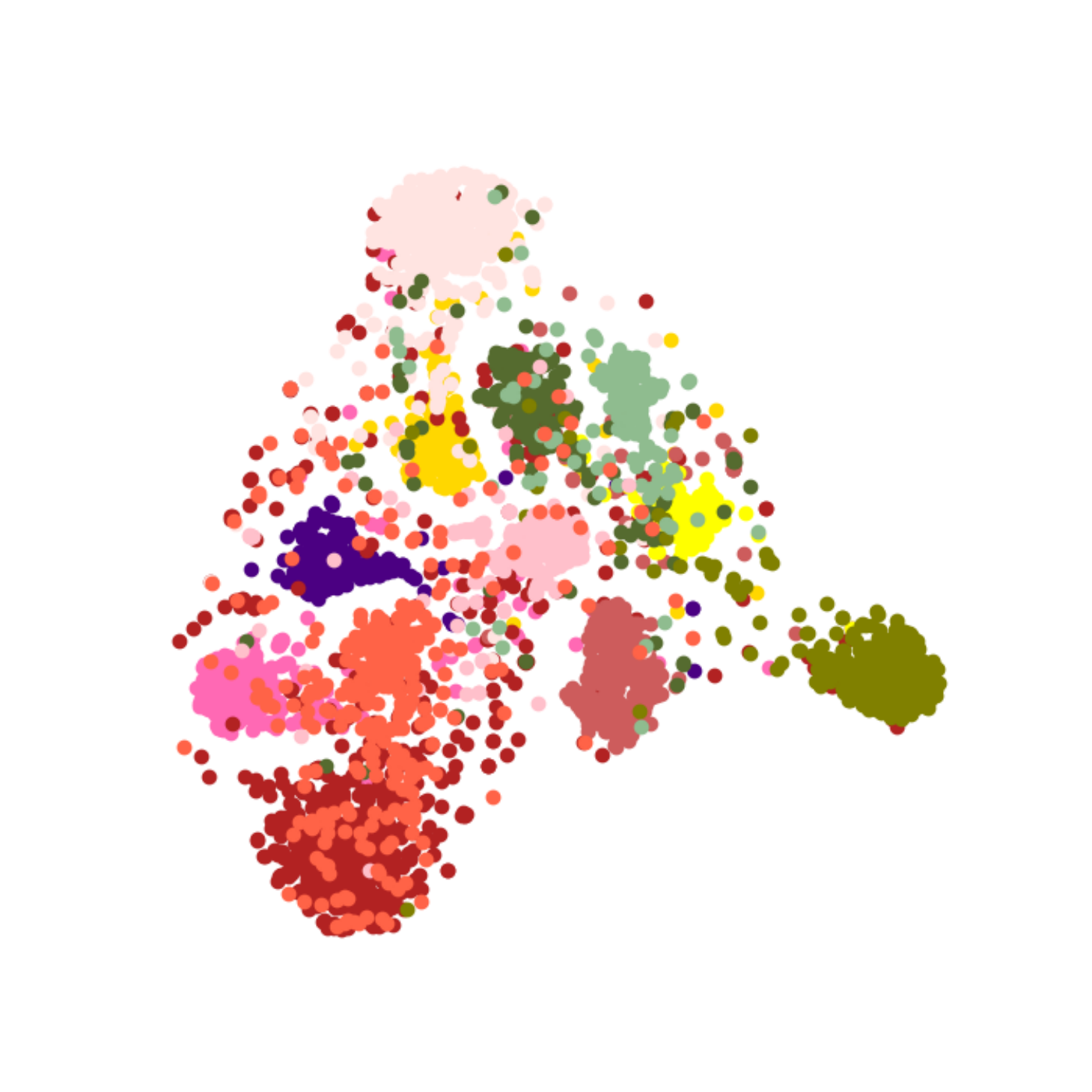}\label{fig:plot_adapted}\vspace{-2mm}}
\end{minipage}
\vspace{-2mm}
\caption{\small Visualization of VisDA-classification (12 classes) target features using T-SNE~(\cite{maaten2008visualizing}): {\bf \subref{fig:plot_original}} features obtained by the Imagenet-pretrained (\cite{deng2009imagenet}) ResNext model not finetuned on VisDA; {\bf \subref{fig:plot_nonadapted}} features from the ResNext model fine-tuned only on VisDA source samples without any adaptation; {\bf \subref{fig:plot_adapted}} features obtained by ResNext adapted by our ADR method.}
\label{fig:embedding}
\end{figure}

Fig.~\ref{fig:embedding} visualizes the target features obtained by $G$ with the pretrained model, model fine-tuned on source, and our ADR method. While the embedding of the source only model does not separate classes well due to domain shift, we can see clearly improved separation with ADR. 

%To find hyper-parameter, we cannot use test images because the labels are completely absent. We tuned hyper-parameters based on validation data. We show the result trained on pretrained ResNet101~(\cite{he2016deep}) and ResNext~(\cite{xie2016aggregated}) models. For feature extraction networks, we use pretrained model whose top fully connected layer is removed. For classification networks, we use three-layered fully connected networks. 

\begin{table}[t]
  {  \tabcolsep=0.8mm
    \begin{center}
          \scalebox{0.9}{
\footnotesize
\begin{tabular}{l|cccccccccccc|c}
\toprule
Method&  \rotatebox{90}{{\small aeroplane}}&\rotatebox{90}{{\small bicycle}}&\rotatebox{90}{{\small bus}}&\rotatebox{90}{{\small car}}&\rotatebox{90}{{\small  horse}}&\rotatebox{90}{{\small knife}}&\rotatebox{90}{{\small motorcicyle}}&\rotatebox{90}{{\small person}}&\rotatebox{90}{{\small plant}}&\rotatebox{90}{{\small skateboard}}&\rotatebox{90}{{\small train}}&\rotatebox{90}{{\small truck}}&\rotatebox{90}{{\small mean}}\\\hline
\multicolumn{2}{c}{{\bf Finetuning on ResNet101}}\\\hline
{\footnotesize Source Only} &55.1&53.3&61.9&59.1&80.6&17.9&79.7&31.2&81.0&26.5&73.5&8.5&52.4\\
{\footnotesize MMD}&87.1&63.0&76.5&42.0&90.3&42.9&85.9&53.1&49.7&36.3&85.8&20.7&61.1\\
{\footnotesize DANN}&81.9&77.7&82.8&44.3&81.2&29.5&65.1&28.6&51.9&54.6&82.8&7.8&57.4\\
{\footnotesize ENT} & 80.3&75.5&75.8&48.3&77.9&27.3&69.7&40.2&46.5&46.6&79.3&16.0&57.0\\
{\footnotesize Ours}  &85.1&77.3&78.9&64.0&91.6&52.7&{\bf 91.3}&{\bf 75.8}&{\bf 88.5}&{\bf 60.5}&{\bf 87.7}&{\bf 33.9}&73.9\\
{\footnotesize Ours (retrain classifer)}  &{\bf 87.8}&{\bf 79.5}&{\bf 83.7}&{\bf 65.3}&{\bf 92.3}&{\bf 61.8}&88.9&73.2&87.8&60.0&85.5&32.3&{\bf 74.8}\\\hline
\multicolumn{2}{c}{{\bf Finetuning on ResNeXt}}\\\hline
{\footnotesize Source Only} &74.3&37.6&61.8&68.2&59.5&10.7&81.4&12.8&61.6&26.0&70.0&5.6&47.4\\
{\footnotesize MMD}&90.7&51.1&64.8&65.6&89.9&46.5&91.9&40.1&81.5&24.1&{\bf 90.0}&28.5&63.7\\
{\footnotesize DANN}&86.0&66.3&60.8&56.0&79.8&53.7&82.3&25.2&58.2&31.0&89.3&26.1&59.6\\
{\footnotesize ENT} & {\bf 94.7} & 81.0 & 57.0 & 46.6 & 73.9 & 49.0 & 69.2 & 31.0 & 40.5 & 34.3 & 87.3 & 15.1 & 56.6\\
{\footnotesize Ours}  &93.5&81.1&82.2&68.3&92.0&67.1&90.8&{\bf 80.1}&{\bf 92.6}&{\bf 72.1}&87.6&42.0&79.1\\
{\footnotesize Ours (retrain classifer)}  &94.6&{\bf 82.6}&{\bf 84.4}&{\bf 73.3}&{\bf 94.2}&{\bf 83.4}&{\bf 92.2}&76.1&91.5&55.1&85.2&{\bf 43.2}&{\bf 79.6}\\
 \hline
  \end{tabular}}
  \vspace{-2mm}
    \caption{{\small Results on Visda2017 classification datasets. DANN and MMD are distribution alignment methods proposed by (\cite{ganin2014unsupervised}) and (\cite{long2015learning}) respectively. Ours (retrain classifier) means the classifier retrained for our proposed generator as we mentioned in Sec \ref{insight}. Our proposed method shows much better performance than existing methods.} }
    \label{table:exp_class}
      \end{center}
  }
\vspace{-3mm} 
\end{table}

\noindent {\bf Image Segmentation experiments.}
Next, we apply our method to adaptation for semantic image segmentation. Image segmentation is different from classification in that we classify each pixel in the image. To evaluate the performance on segmentation, the synthetic GTA5~(\cite{richter2016playing}) dataset is used as source, and real CityScape~(\cite{cordts2016cityscapes}) dataset is used as target. Previous work tackled this problem by matching distributions of each pixel's feature in a middle layer of the network~(\cite{hoffman2016fcns}).
In this work, we apply ADR by calculating sensitivity between all pixels. The training procedure is exactly the same as in classification experiments. 

We use the pretrained ResNet50, and utilize an FCN~(\cite{long2015fully}) based network architecture. For the feature generator, we use the pretrained network without fully-connected layers. For the classifier, we use a fully-convolutional network with dropout layers. Due to limited memory, the batch size is set to 1. We include details of the network architecture in our supplementary material. For comparison, we train a domain classifier based model for our network (DANN). We build a domain classifier network for the features of each pixel following~(\cite{hoffman2016fcns}).

\begin{figure}[t]
      \vspace{-2mm}
  \begin{minipage}{0.49\hsize}

  \centering
  \includegraphics[width=\hsize]{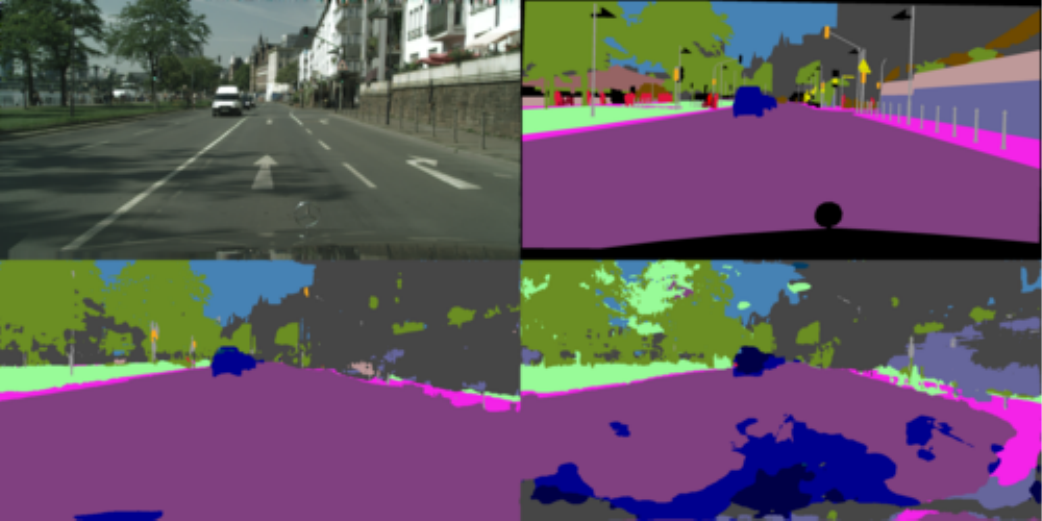}
\end{minipage}
\begin{minipage}{0.49\hsize}
  \centering
    \includegraphics[width=\hsize]{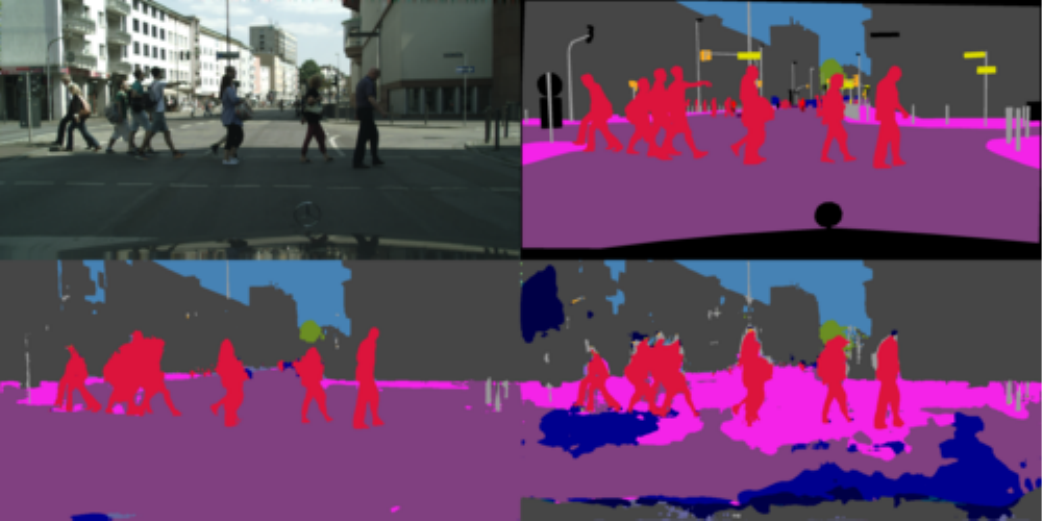}
\end{minipage}
\vspace{-2mm}
\caption{{\small Comparison of results on two real images. Clockwise from upper left: Original image; Ground truth; Segmented image before adaptation; Segmented image after adaptation by our method.}}
\label{fig:seg_image}
\vspace{-4mm}
\end{figure}

In Table~\ref{table:exp_seg}, we show the qualitative comparison with existing methods. ADR clearly improves mean IoU compared to the source-only and competing models. We illustrate the improvement on example input images, ground truth labels, images segmented by Source Only model and our method in Fig. \ref{fig:seg_image}. While the Source Only model seems to suffer from domain shift, ADR generates a clean segmentation. These experiments demonstrate the effectiveness of ADR on semantic segmentation. Although we implemented ENT in this setting, the accuracy was much worse than the Source Only model with a mIoU of 15.0. The ENT method does not seem to work well on synthetic-to-real shifts.

\begin{center}
\begin{table}[t]
  {  \tabcolsep=0.5mm
    \small
  \begin{tabular}{l|c|c|c|c|c|c|c|c|c|c|c|c|c|c|c|c|c|c|c|c|c|c}
\toprule
Method&  \rotatebox{90}{{\small road}}&\rotatebox{90}{{\small sidewalk}}&\rotatebox{90}{{\small building}}&\rotatebox{90}{{\small wall}}&\rotatebox{90}{{\small  fence}}&\rotatebox{90}{{\small pole}}&\rotatebox{90}{{\small t light}}&\rotatebox{90}{{\small t sign}}&\rotatebox{90}{{\small veg}}&\rotatebox{90}{{\small terrain}}&\rotatebox{90}{{\small sky}}&\rotatebox{90}{{\small person}}&\rotatebox{90}{{\small rider}}&\rotatebox{90}{{\small car}}&\rotatebox{90}{{\small truck}}&\rotatebox{90}{{\small bus}}&\rotatebox{90}{{\small train}}&\rotatebox{90}{{\small mbike}}&\rotatebox{90}{{\small bike}}&{{\small mIoU}}\\\hline
{\footnotesize Source Only} & 64.5 & 24.9 & 73.7 & 14.8 & 2.5 & 18.0 & 15.9 & 0 & 74.9 & 16.4 & 72.0 & 42.3 & 0.0 & 39.5 & 8.6 & 13.4 & 0.0 & 0.0 & 0.0 & 25.3\\\hline
{\footnotesize DANN}&72.4&19.1&73.0&3.9&9.3&17.3&13.1&5.5&71.0&20.1&62.2&32.6&5.2&68.4&12.1&9.9&0.0&5.8&0.0&26.4\\\hline
{\footnotesize FCN Wild}& 70.4 & {\bf 32.4} & 62.1 & 14.9 & 5.4 & 10.9 & 14.2 & 2.7 & 79.2 & 21.3 & 64.6 &{\bf 44.1} & 4.2 & 70.4 & 8.0 &7.3 & 0.0 & 3.5 &0.0& 27.1\\\hline
{\footnotesize Ours}  & {\bf 87.8} & 15.6 & {\bf 77.4} & {\bf 20.6} &{\bf 9.7} & {\bf 19.0} & {\bf 19.9} &{\bf 7.7} &{\bf 82.0} &{\bf 31.5} &{\bf 74.3} & 43.5 &{\bf 9.0} &{\bf 77.8} &{\bf 17.5} &{\bf 27.7} &{\bf 1.8} &{\bf 9.7} &0.0& {\bf 33.3}\\
 \hline
  \end{tabular}
  \vspace{-2mm}
    \caption{{\small Results on adaptation from GTA5 $\rightarrow$ Cityscapes. DANN and FCN Wild denote methods proposed by (\cite{ganin2014unsupervised}) and (\cite{hoffman2016fcns} respectively.}}
    \label{table:exp_seg}
    }
\end{table}
\end{center}

\vspace{-8mm}
\subsection{Semi-supervised learning using GANs}
      
In this section, we demonstrate the effectiveness of our method in training a Generative Adversarial Network (GAN) applied to semi-supervised learning.
We follow the method proposed by~(\cite{springenberg2015unsupervised,salimans2016improved}), who use a $K$-class classification network as a critic to train a GAN in the semi-supervised setting.

\noindent {\bf Approach.} In contrast to the domain adaptation setting, here $G$ tries to generate images which fool the critic $C$. Also, in this setting, we are given labeled and unlabeled real images from the same domain. Then, we train the critic to classify labeled  images correctly and to move unlabeled images far from the decision boundary. To achieve this, we propose to train the critic with the following objective:
\vspace{-3mm}
\begin{equation}
    \mymin_{C} L_{C} = L(X_{L},Y_{L})+L_{adv}(X_{u})- L_{adv}(X_{g}) + {\mathbb{E}_{\mathbf{x_{u}}\sim X_{u}}}\sum_{k=1}^{K}p(y=k|{\mathbf x_{u}}) \log p(y|{\mathbf x_{u}})\\
  \end{equation}
\begin{equation}
  L_{adv}(X_{u}) = {\mathbb{E}_{\mathbf{x_{u}}\sim X_{u}}}[d(C_1(G(\mathbf{x_{u}})),C_2(G(\mathbf{x_{u}})))]
\end{equation}
\begin{equation}
  L_{adv}(X_{g}) = {\mathbb{E}_{\mathbf{x_{g}}\sim X_{G}}}[d(C_1(G(\mathbf{x_{g}})),C_2(G(\mathbf{x_{g}})))]
  \end{equation}
where $X_{L}$ denotes the subset of labeled samples, $X_{u}$ denotes unlabeled ones and $X_{g}$ denotes  images generated by $G$. The critic is trained to minimize the loss on labeled samples in the first term. Since unlabeled  images should be far away from the decision boundary and  should be distributed uniformly among the classes, we add the second and fourth term. The third term encourages the critic to detect fake images generated near the boundary. 

The objective of $G$ is as follows, 
\begin{equation}
  \mymin_{G} L_{adv}(X_{g}) + ||{\mathbb{E}_{x_{g}\sim X_{g}}f(\mathbf{x_{g}})}-{\mathbb{E}_{\mathbf{x_{u}}\sim X_{u}}f(\mathbf{x_{u}})}||^2
  \end{equation}
where the second term encourages generated images to be similar to real images, which is known to be effective to stabilize the training. The first term encourages the generator to create fake images which should be placed far away from the boundary. Such images should be similar to real images because they are likely to be assigned to some class with high probability. 
%Using these objectives, we trained $C$ and $G$ in an adversarial way. Unlike in domain adaptation experiments, 
Here, we update $C$ and $G$ same number of times. 

\noindent {\bf Experiment.} We evaluate our proposed GAN training method by using SVHN and CIFAR10 datasets, using the critic network architecture from~(\cite{salimans2016improved}). In the experiment on SVHN, we replaced Weight Normalization with Batch Normalization for $C$. Also, in the experiment on CIFAR10, we construct a classifier from a middle layer of the critic, which is not incorporated into the adversarial training step. This is motivated by the insight that the critic in our method is trained to be too sensitive to the dropout noise as we explained in Sec~\ref{insight}. 
%With this modification, we could not see clear improvement for SVHN. 

\begin{figure}[t]
\begin{minipage}{0.49\hsize}
  \centering
  \subfigure[SVHN]{\includegraphics[width=\hsize]{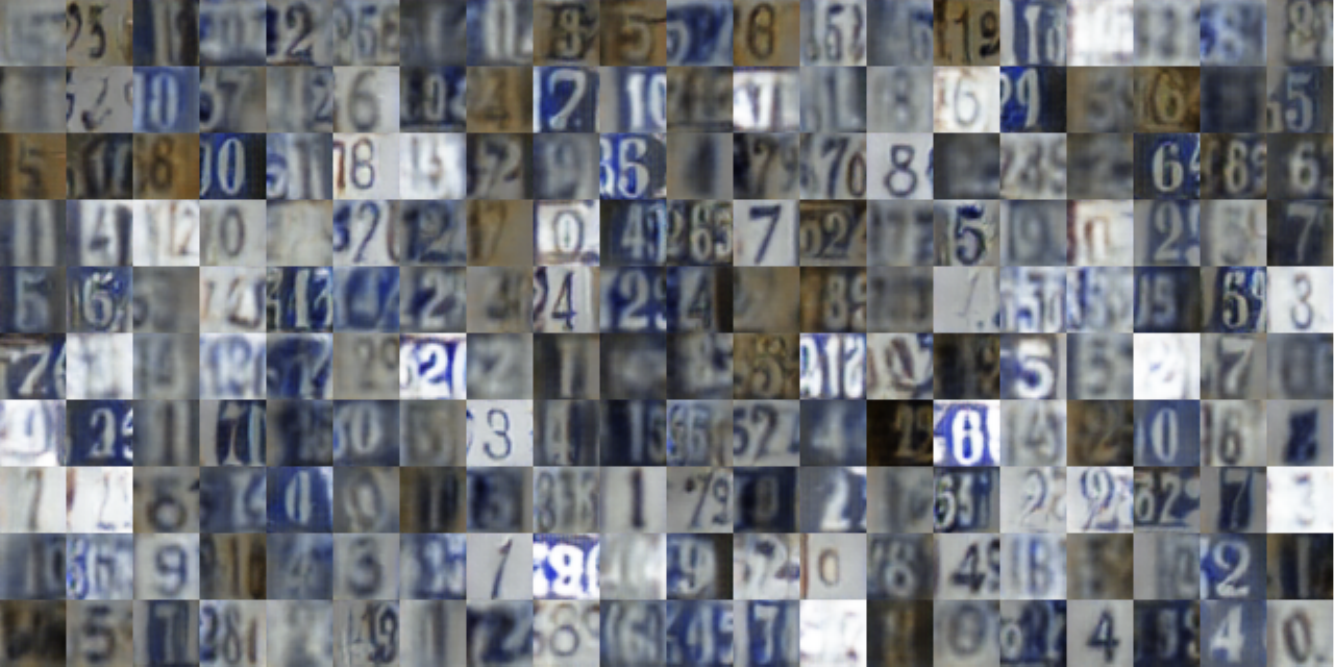}\label{fig:svhn}}
\end{minipage}
\begin{minipage}{0.49\hsize}
  \centering
  \subfigure[CIFAR10 ]{\includegraphics[width=\hsize]{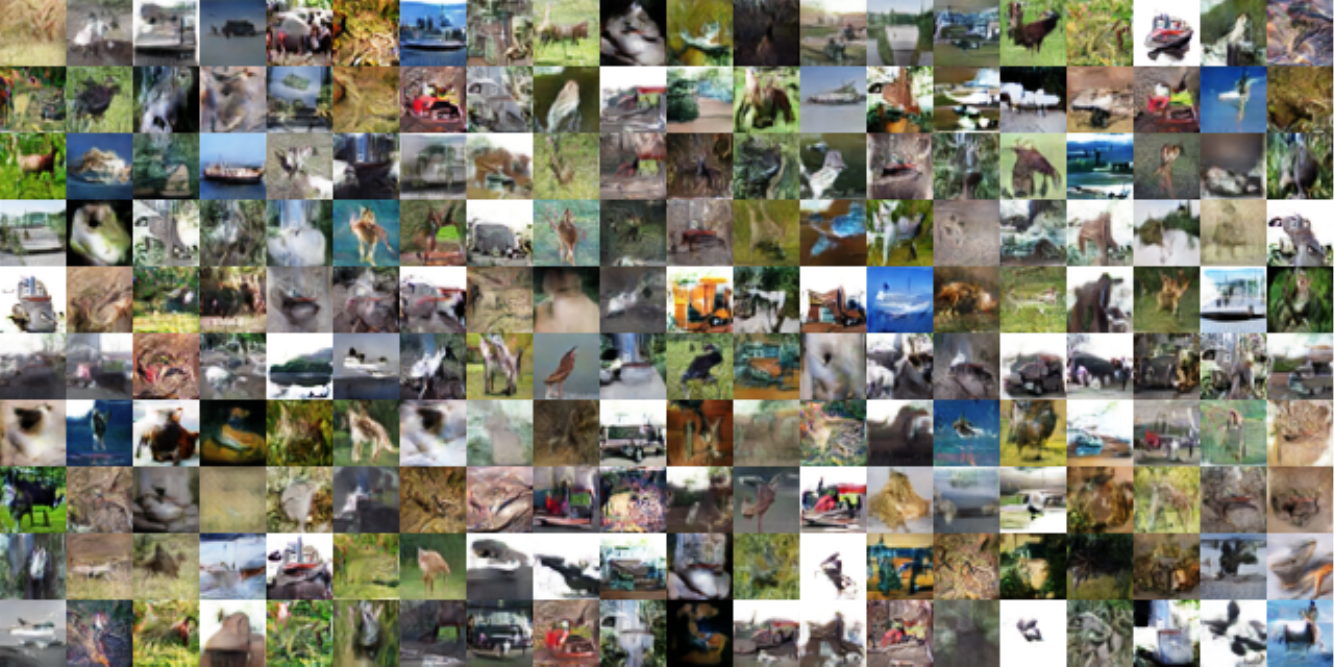}\label{fig:cifar}}
\end{minipage}
\vspace{-2mm}
\caption{Examples of generated images. }
\label{fig:gan_image}
\end{figure}

\noindent {\bf Results.} From Fig. \ref{fig:svhn}, we can see that ADR seems to generate realistic SVHN images. Some images are significantly blurred, but most of images are clear and diverse.  As for generated CIFAR10 images, they do not seem as realistic, but some objects appear in most images.
In Table~ \ref{table:exp_gan}, we can see that the accuracy of the critic trained by our method has better performance than other models for SVHN. For CIFAR10, the accuracy was comparable to other state-of-the-art methods. This demonstrates that our proposed ADR approach is effective for training semi-supervised GANs. 

\begin{table}[t]
\begin{center}
  \begin{tabular}{lcccccc}
\toprule
& SVHN (\% errors) &&CIFAR (\% errors)\\
Labeled Only &&&&\\\hline
SDGM~(\cite{maaloe2016auxiliary}&16.61 $\pm$ 0.24&&-&&\\
CatGAN~(\cite{springenberg2015unsupervised})&- & &19.58$\pm$0.46&&\\
ALI~(\cite{dumoulin2016adversarially})&7.42$\pm$0.65&&{\bf 17.99}$\pm$1.62\\
  ImpGAN~(\cite{salimans2016improved})&8.11$\pm$1.3 & &18.63$\pm$2.32&&\\\hline

 Ours  &{\bf 6.26}$\pm$1.05&&19.63$\pm$0.37\\
 \hline
  \end{tabular}
    \caption{Comparison with state-of-the-art methods on two benchmark datasets. Only methods without data augmentation are included. We used the same critic architecture as used in ImpGAN.}
\label{table:exp_gan}
\end{center}
\end{table}

\vspace{-2mm}        
\section{Conclusion}
\vspace{-3mm}        
In this paper, we introduced a novel approach for aligning feature distributions called Adversarial Dropout Regularization, which learns to generate discriminative features for the target domain. The method consists of a critic network that can detect  samples near the boundary and a feature generator that fools the critic. Our approach is general, applies to a variety of tasks, and does not require domain labels.
%We used dropout to detect the samples near the boundary and trained networks in adversarial way. 
In extensive domain adaptation experiments, our method outperformed baseline methods, including entropy minimization, and achieved state-of-the-art results on three datasets. It also proved effective when applied to  Generative Adversarial Networks for semi-supervised learning. 
%Future work is to investigate a theoretical aspect of our method.
\section{Acknowledgements}
We would like to thank Trevor Darrel for his great advice on our paper. First author's stay at Boston University is partially supported by scholarship of the University of Tokyo.
The work was partially funded by the ImPACT Program of the Council for Science, Technology, and Innovation
(Cabinet Office, Government of Japan), and was partially supported by CREST, JST. Saenko was supported by IARPA and NSF grants CCF-1723379 and IIS-1724237.

\bibliography{arxiv}
\bibliographystyle{iclr2018_conference}
\appendix

\section*{Appendix}
\vspace{-3mm}
\section{Entropy Based Method for Domain Adaptation}
\label{append:ent}
\vspace{-3mm}
Our method aims to move target samples away from the decision boundary. Some techniques used in training Generative Adversarial Networks can be applied to achieve our goal too.~(\cite{springenberg2015unsupervised,salimans2016improved}) used small number of labeled samples to train critic. Critic is trained to classify real samples into $K$ classes. They also trained critic to move unlabeled real images away from the boundary by minimizing entropy of the critic's output. Generated fake images are moved near the boundary by maximizing the entropy. On the other hand, generator is trained to generate fake images which should be placed away from the boundary.
This kind of method can be easily applied to domain adaptation problem. We would like to describe the method along with our problem setting.

Similar to our method, we have critic networks $C$ and generator $G$. $C$ classifies samples into $K$ class.
$C$ is trained to maximize the entropy of target samples, which encourages to move the target samples near the boundary. Then, $G$ is trained to minimize the entropy of them. Thus, $G$ tries to move target samples away from the boundary.

The only difference from our method is that we used entropy term for adversarial training loss. That is, in this method, we replace our sensitivity term $d(p_1,p_2)$ in Eq. \ref{eq:sensitivity} with entropy of the classifier output. The adversarial loss for this baseline method is a following one.
\begin{eqnarray}
  L_{adv}(X_{t}) &=&{\mathbb{E}_{\mathbf{x_{t}}\sim X_{t}}}[H[p(\mathbf{y}|\mathbf{x_t})]\\
    \label{eq:adv_entropy}
  H[p(\mathbf{y}|\mathbf{x_t})] &=& - \sum_{k=1}^{K}p(y=k|\mathbf{x_t})\log p(y=k|\mathbf{x_t})
    \label{eq:entropy}
\end{eqnarray}

The hyperparameter $n$, how many times we update $G$ for adversarial loss in one mini-batch, is set as $n=4$. Experimentally, it worked well for all settings.  

%In our understanding, this is the same as distribution matching in terms of decision boundary space.
%Maximizing the entropy of target samples means that we try to move the target samples to the area where they are not assigned to any real classes. This method tries to distinguish target samples from source samples because source samples are assigned to some real classes. In this respect, this is similar to making a new class $K+1$ and assigning target samples to it. Similar discussion was made in~(\cite{salimans2016improved}). They discussed that critic does not have to classify fake images into $K+1$ class. $1$$\sim$$K$ indicate each class of real images. $K+1$ is the class for fake images. 
%Critic classifying $K$ class is enough to move the fake samples away from real images by using entropy maximization.
%However, the problem of applying this method to domain adaptation is that this method aims to completely match distributions between source and target in terms of final output space. Therefore, this method is expected to fail in adaptation between distinct domains. We experimented this method in digits classification, object classification and segmentation experiment. In segmentation experiment, this method does not work well at all.
\vspace{-3mm}
\section{Digits Classification Training Detail}
\vspace{-3mm}
We follow the protocol used in~(\cite{tzeng2017adversarial}). For adaptation from SVHN to MNIST, we used standard training splits of each datasets as training data. For evaluation, we used test splits of MNIST. For the adaptation between MNIST and USPS, we sampled 2000 images from MNIST and 1800 images from USPS.
In these experiments, we composed the mini-batch half from source and half from target samples. The batchsize was set as 128 for both source and target. We report the score after repeating Step 1$\sim$3 (please see Sec \ref{sec:procedure}) 20000 times. %parameter updates in this experiment.
For our baseline, we used exactly the same network architecture.
\vspace{-3mm}
\section{Object Classification Training Detail}
\vspace{-3mm}
In this experiment, SGD with learning rate $1.0\times10^{-3}$ is used to optimize the parameters. 
For the finetuning of ResNet101, we set batch-size as 32. Due to the limit of GPU memory, we set it as 24 in finetuning ResNext model. We report the score after 20 epochs training.
In order to train MMD model, we use 5 RBF kernels with the following standard deviation parameters:
\begin{equation}
  \sigma = [0.1,0.05,0.01,0.0001,0.00001]
\end{equation}
We changed the number of the kernels and their parameters, but we could not observe significant performance difference. We report the performance after 5 epochs. We could not see any improvement after the epoch.

To train a model~(\cite{ganin2014unsupervised}), we used two-layered domain classification networks. Experimentally, we did not see any improvement when the network architecture is changed. According to the original method~(\cite{ganin2014unsupervised}), learning rate is decreased every iteration. However, in our experiment, we could not see improvement, thus, we fixed learning rate $1.0\times10^{-3}$. We report the accuracy after 1 epoch. The accuracy dropped significantly after the first epoch. We assume this is due to the large domain difference between synthetic and real images.

For our new baseline, ENT, we used the same hyperparameter as we used for our proposed method. %Although our proposed method continues to increase its accuracy and stably converge,
Since the accuracy of ENT drops significantly after around 5 epochs, we report the accuracy after 5 epoch updates.
\vspace{-3mm}
\section{Segmentation Experiments Detail}
\vspace{-3mm}
\begin{wrapfigure}[11]{r}[5mm]{60mm}
  \begin{center}
    \vspace{-16mm}
  \includegraphics[width=50mm]{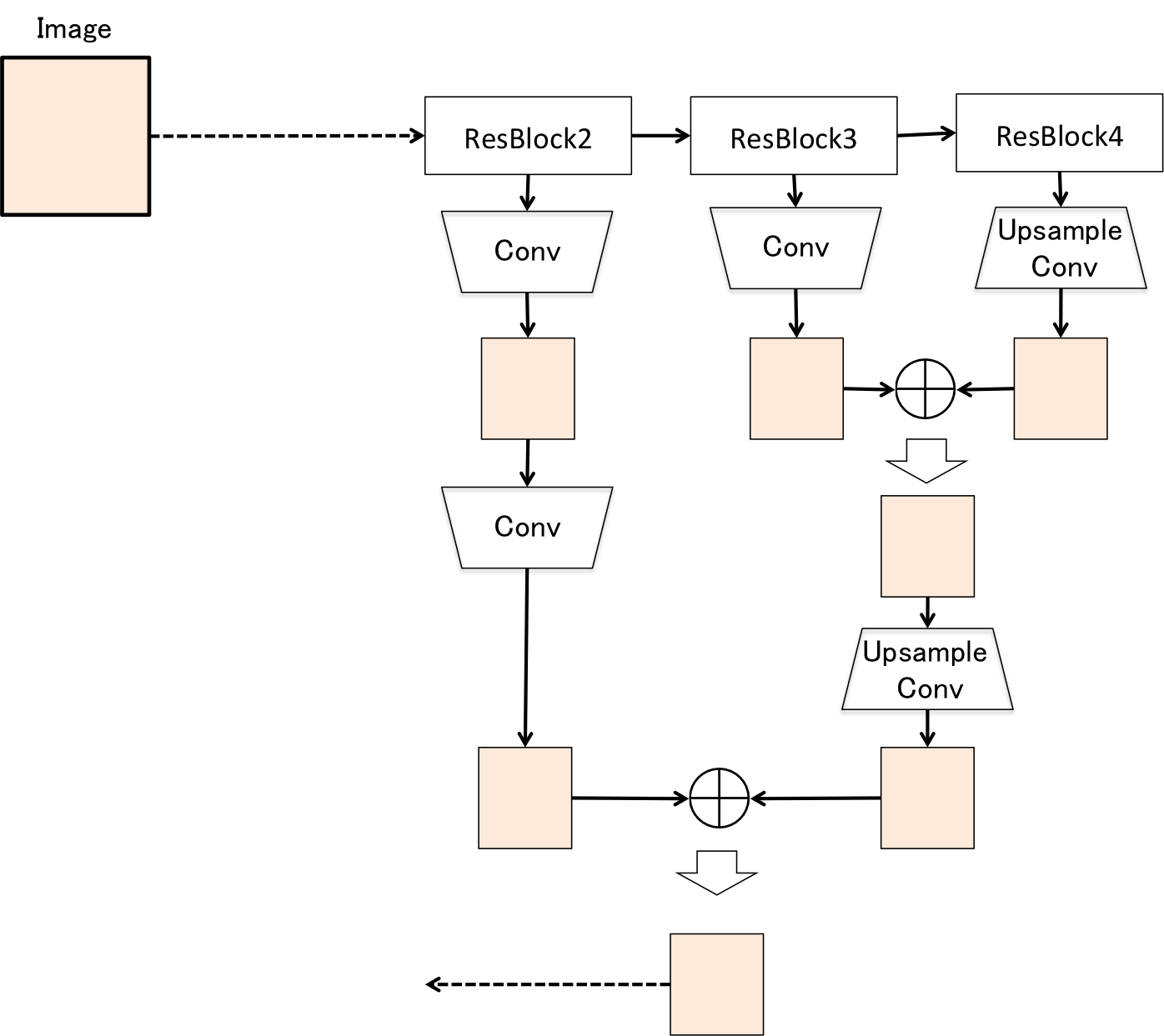}
  \end{center}
        \vspace{-4mm}
  \caption{Overview of architecture for semantic segmentation}

  \label{fig:architecture}
  \end{wrapfigure}
We modified FCN \cite{long2015fully} architecture suitable for ResNet structure. The features from ResBlock 2$\sim$4 and the first convolution layer and maxpooling layer are used in our implementation. In Fig. \ref{fig:architecture}, we show how we integrated the features of each layers. We regard the layers of ResNet50 as generator and rest of the networks, namely convolution and upsampling layers as a critic network. The input images were resized to 512x1024 due to the limit of GPU memory. For the same reason, the batchsize was set to one. 
\vspace{-3mm}
\section{GAN training}
\vspace{-3mm}
We used the critic architecture proposed by \cite{salimans2016improved}. We set the batch size as 100 and used Adam with learning rate $2.0\times1.0^{-4}$ for optimizer. 
After the conv6 layer of the critic, we constructed a classifier which was not concerned with adversarial learning process. 

\end{document}